# Quantifying structure differences in literature using symbolic diversity and entropy criteria

Gerardo Febres[1,2] Klaus Jaffé[2]
gerardofebres.usb.ve kjaffe@usb.ve

[1] Departamento de Procesos y Sistemas, Universidad Simón Bolívar, Venezuela.
[2] Laboratorio de Evolución, Universidad Simón Bolívar, Venezuela.

*ABSTRACT. We measured entropy and symbolic diversity of texts written in English and Spanish. We included texts by Literature Nobel laureates and other famous authors. We formed four groups of texts according to the combinations of language used and the author's Literature Nobel Prize condition. Entropy, symbol diversity and symbol frequency profiles were compared for these four groups. We also built a scale sensitive to the quality of writing and evaluated its relationship with the Flesch´s readability index for English and the* Szigriszt´s perspicuity index *for Spanish. Results suggest a correlation between entropy and word diversity with quality of writing. Text genre also influences the resulting entropy and diversity of the text. Results suggest the plausibility of automated quality assessment of texts.*

## 1 Introduction

In 1880 Lucius Sherman (Sherman, 1893) (DuBay, 2004) studied the structure of the English language from a statistical point of view, finding that the average number of words in English sentences had diminished from 45, at the times of Queen Elizabeth I, to 23 during the time Sherman lived. A second result showed that writers are consistent in the average number of words per sentence (DuBay, 2004). Efforts to construct methods to evaluate text readability have continued since then. During the early twentieth century, teachers evaluated texts relying on the *Teacher's Word Book* (Thorndike, A Teacher's Word Book, 1921) by Thorndike; a collection of the 10,000 most frequently used words in English published in 1921 and extended to 20,000 words in 1932 (Thorndike, A Teacher's Word Book of 20,000 Words, 1932) and 30,000 in 1994 (Thorndike, The Teacher's Word Book of 30,000 Words, 1944). These word-frequency lists were mostly used to evaluate the appropriateness of reading material for children at elementary schools. The evaluation of quality of writing consisted, basically, in counting the number of different words in a text as a measurement of the author's size of vocabulary.

The vocabulary lists became the basis for describing an underlying structure, as is the English language word frequency distribution, known today as the Zipf's law (Kirby, 1985). Due to George Kingsley Zipf's renowned work, Human Behavior and The Principle of Least Effort (Zipf, 1949). The evaluation of quality of writing consisted, basically, in counting the number of different words in a text as a measurement of the author's size of vocabulary.

Starting with his PhD thesis (Flesch, Marks of a readable style, 1943), Rudolf Flesch published a series of books studying English texts: (Flesch, The art of plain talk, 1946) (Flesch, The art of readable writing, 1949) (Flesch, How to test readability, 1951) (Flesch, The art of clear thinking, 1951, 1973) (Flesch, How to write, speak and think more effectively, 1958). These efforts led to the Reading Ease Score, usually referred to as $RES$, a formula based on the weighted combination of vocabulary, average word character-length and average sentence word-length, useful to evaluate the ease, or difficulty, to read and understand the content of texts. After Flesch's original work, other researchers built formulas based on to Flesch's $RES$ formula. Adaptations for specific uses such as the evaluation of applicants to enter the US navy (Kincaid, Fishbourne, Rogers, & Chissom, 1975) and institutions in charge of assessing reading and comprehension of prospective students of American universities (Kathleen M. Sheehan, Irene Kostin, Yoko Futagi, Michael Flor, 2010), as well as for analyzing the suitability of basic school texts appeared and became the theme of much research and experimentation. Within the fields where readability formulas have been a useful tool, health occupies an important place (Barrio Cantalejo, 2008) (Trauzettel-Klosinski, Susanne; Dietz, Klaus; Group, the IReST Study, August 2012) (Gröne), but the field of education resulted best suited for the application of these readability formulas versions specially made by (Chall, 1958), Kincaid and others.

(Fernández-Huerta, 1959) adjusted the original Flesch readability formula and produced the 'Formula de lecturabilidad de Fernández-Huerta' (Readability Formula) for Spanish. Another adaptation of the Flesch's formula, presented by (Szigriszt-Pazos, 1993) named 'Formula de Perspicuidad' (Perspicuity Formula) or $IPSZ$, as we will refer to it, has become the current standard to evaluate the readability of Spanish texts.

In recent years, a different approach to measure readability has appeared. Relying on today's computing capacity, (Tanaka-Ishii, Tezuka, & Terada, 2010), proposed looking at readability as a relative property of texts instead of an absolute assessment. Theirs is a method based on Support Vector Machines and sorting algorithms. Yet, traditional readability formulas are widely accepted, and remain as the most used method to evaluate the appropriateness of texts in accordance with the audience they are intended for.

The relationship between readability measures and word frequency profiles is the focus during the 1960's by (Klare, 1968). Klare added to the 'Human Behavior and The Principle of Least Effort', the mechanisms that explain the high frequency of appearance short words in natural language texts. Klare states that the size of words is an underlying 'learning' factor which makes the communication process more effective, since shorter words are faster and better understood by both interacting parts, the emitter and the receiver, highlighting the fact that any communication process is not only less laborious, but also more effective when shorter symbols are used.

The possibility of measuring the quality of writing became a need with the emergence of qualifying tests for schools and colleges. Writing skills, in spite of being a neat reflex of intellectual capabilities, are an elusive property to measure. On the other hand, complexity measurements of text messages, address only the evaluation of the quantity of information needed to specify and transmit a message, compressibility and other aspects of the information, focusing on the mere descriptive process and disregarding the idea content, beauty or any other form of valuable characteristic of the message. As an alternative, we suggest evaluating quality of writing by formulas based on characterizations of texts Zipf's profiles. The particular language's grammar rules establish restrictions over some degrees of freedom of the symbol frequency distribution profile (Febres, Jaffé, & Gershenson, 2014), but there is still enough space for the text's symbol frequency profile, to be sensitive to some properties of a text as for example: organization, coherence, vocabulary richness, length of sentences and word difficulty, which have influence over readability.

Entropy was suggested as an index sensitive to the writing style by (Kontoyiannis, 1997). In his study, Kontoyiannis computed entropy at the scale of characters; in other words, entropy was estimated considering a fixed symbolic diversity determined by the 26 characters of the English alphabet. Despite the obvious weak connection between the letter frequency distribution and the style of writing in any natural language, Kontoyiannis was able to conjeture the existence of some correlation between entropy and the style of writing. In another path of research, Jackes Savoy (Savoy, Text Clustering: An Application with the State of the Union Addresses, in press) (Savoy, Vocabulary Growth Study: An Example with the State of the Union Addresses, in press) presents evidence of the influence of the time period and the political affiliation of the authors and the frequency of use of specific type-of-words as verbs, pronouns, and adverbs. Savoy used a sample of a few hundred speeches pronounced by American presidents.

In this paper we investigate the impact of quality of writing over word diversity, entropy and ranked frequency profiles. To perform our experiments, we built a library with 138 English texts and 136 Spanish texts. The authors of the texts include politicians, military, Literature Nobel laureates, writers, scientists, artists, and other public figures. To overcome the bias introduced by the variety of text lengths, we evaluated differences in symbol diversity and entropy indices that might be related to writing quality. Finally, we propose an evaluation scale for English and Spanish that, we claim, is related to the quality or writing. Representing the set of speeches in the plane specific diversity-entropy, we visually highlight that relationship.

## 2 Methods

We based this work on a library containing English texts and Spanish texts. Texts were grouped in two categories: one integrated by those texts originated by authors who were laureate with the Literature Nobel Prize, the other formed by texts produced by renowned writers, politicians, military and social personalities. Combining writers and Nobel laureates for English and Spanish, we obtained four groups for our analysis.

Each text was characterized by its symbolic diversity D, entropy h, and distribution of symbolic frequency f in accordance with the definitions shown below. We built a mathematical model with these properties for the four groups created. Mean values and dispersion were studied by statistical methods. Finally we produced quality of writing scales for English and Spanish.

### 2.1 Text length $L$ and symbolic diversity $d$

The length of a text $L$ is measured as the total number of symbols used, and the diversity $D$ as the number of different symbols that appear in the text. We define the specific diversity $d$ as the ratio of diversity $D$ and length $L$ , that is

$$d = \text{specific diversity} = D/L \, . \qquad (1)$$

As symbols we consider words as well as punctuation signs, therefore the number of symbols is obtained adding the count of both types of symbols.

### 2.2 Entropy $h$

Shannon's entropy expression (Shannon, 1948) is used to measure texts information. Symbols (words) are treated as information units, disregarding any differential information weight that may be associated to the word meanings, length or context. The entropy $h$ for texts is evaluated following definition:

$$h = -\sum_{r=1}^{D} \frac{f_r}{L} \, log_D \, \frac{f_r}{L} . \tag{2}$$

where $f_r$ is the number of appearances of the symbol occupying the place r within the ranking of symbols' frequency. Notice the base of the logarithm is the diversity $D$ and hence $h(L, D)$ is bounded between zero and one. Setting the base of the logarithm to 2, expression (2) becomes

$$h = -\frac{1}{log_2 D} \sum_{r=1}^{D} \frac{f_r}{L} \, log_2 \, \frac{f_r}{L} . \tag{3}$$

## 2.3 Symbol frequency distribution *f*

When the symbols of a message are arranged according to the number of their appearances, from the most frequently found symbol to the least, we obtain the ranked symbol profile. For any symbol profile, the number of words in a rank segment $[a, b]$ may be computed as:

$$L_{a,b} = \sum_{r=a}^{b} f_r . \tag{4}$$

where $r$ is a word frequency rank position while $a$ and $b$ are the start and the end of the considered symbol rank segment respectively. For any segment, a=1 and b=D.

## 2.4. Zipf's Deviation *J*

Zipf's law states that any sufficiently long English text will behave according to the following rule (Kirby, 1985) (Gelbukh & Sidorov, 2001):

$$f(r) = \frac{f_a}{(r-a)^g} , \tag{5}$$

where $r$ is the ranking by number of appearances of a symbol, $f(r)$ a function that retrieves the numbers of appearances of word ranked as $r$, $f_a$ the number of appearances of the first ranked word within the segment considered, and $g$ a positive real exponent.

For any message, we define Zipf's reference $Z_{a,b}$ as the total number of symbol appearances in the ranking segment [a, b] assuming that it follows Zipf's Law. Therefore $Z_{a,b}$ is

$$Z_{a,b} = \sum_{r=\mathrm{a}}^{b} f_r = \sum_{r=a}^{b} \frac{f_a}{r^g} . \tag{6}$$

The complete message Zipf's reference $Z_{1,D}$ is determined by expression (6) and the corresponding Zipf's deviations for the whole distribution $J_{1,D}$ is

$$J_{1,D} = {(L_{1,D} - Z_{1,D})}/{Z_{1,D}} . \tag{7}$$

## 2.5 Model Relative Deviations

As explained in (Grabchak, Zhiyi, & Zhang, 2013), statistics of specific diversity and entropy for natural languages texts have a bias upon the text length. This bias is due to the language structure and the definition of these properties. To compensate for the bias introduced by the diversity on text length of our library, we used a minimal square error regression to model specific diversity and entropy. The difference between the properties from data and the regression model is referred to as *Relative Deviation*. Applied cases to diversity relative deviation $d_{rel}$ and entropy relative deviation $h_{rel}$ are included in Eqs. (8) and (9).

$$d_{rel} = \frac{D - D_m}{D_m} . \tag{8}$$

$$h_{rel} = h - h_m . \tag{9}$$

Notice that Zipf's deviation, calculated as Expression (7) indicates, also works expresses a relative deviation.

## 2.6 Writing Quality Scale $WQS$

We did not find any computerized method to evaluate quality of writing. Thus, we designed a method for evaluating the quality of writing which results in a value we called Writing Quality Scale ($WQS$). Our method is based on evaluations of Equations (7), (8) and (9) for several hundred texts organized in groups as will be explain in Section 2.8.

## 2.7 Readability formulas $RES$ and $IPSZ$

Readability formulas are available for many languages. They do not measure quality of writing but the appropriateness of a text for certain group of readers like, for example, children belonging to a school grade. Thus we used some readability formulas as a reference to compare $the\ WQS$ with them. For English we used the Reading Easy Score ($RES$) by Flesch (Flesch, How to test readability, 1951):

$$RES = 206.835 - 84.6\,W - 1.015\,S , \tag{10}$$

where $W$ is the average of the word length measured in syllables and $S$ the average of the phrase length measured in words. For Spanish, we used the adaptation that Szigriszt (Szigriszt-Pazos, 1993) made to the $RES$ formula, named the Perspicuity Index($IPSZ$):

$$IPSZ = 206.835 - 84.6\,W - S . \tag{11}$$

In Eq. (11) $W$ and $S$ represent the same as in $RES$ formula. Values of $S$ were obtained as $S = L_W/L_{Ph}$ where $L_w$ is the text length measured in words and $L_{Ph}$ is the text length measured in phrases. In English as well as in Spanish, a phrase ends every time a period, colons, semicolons, question mark, exclamation sign or ellipsis appears. Thus $L_{Ph}$ equal the addition of the appearances of the mentioned punctuation signs. The average number of syllables per word $W$ is calculated as

$$W = L_{SY}/L_W , \tag{12}$$

where $L_{SY}$ is the number of syllables of the whole text. Determining the number of syllables L_SY, is more difficult than counting words or punctuation signs; syllables are the textual representation of single sounds, whose start and end may be difficult to recognize, and additionally, the rules to extract syllables from a text have many exceptions, and vary from language to language. In fact, some authors (van den Bosch, Content, Daelemans, & de Gelder, 1994) refer to the deviation from a regular correspondence between a written symbol and the associated phoneme, as *letter-phoneme complexity*, or *orthographic depth*; a completely different notion of complexity from the one we are dealing with in the present study. Thus, recognizing syllables from graphemes with an automated process is not a straight forward task. Especially for English, which is considered an *orthographically deep language*[1], strict correspondence between writing and pronunciation, and vice versa, rarely exists. For Spanish, there is correspondence from writing to pronunciation, meaning that starting from a written word, we know its sound; but there may be many ways of writing down a sound we hear. This ambiguous correspondence between writing and pronunciation appears in English, Spanish and up to some degree in most alphabetic natural languages[2]. To prevent the writing of software codes to count syllables in English and Spanish texts, we decided to estimate $L_{SY}$ by computing

$$L_{SY} = L_{CH}/C_{SY}\,, \tag{13}$$

where $L_{CH}$ is the number of characters not including punctuation signs in the text and $C_{SY}$ is the average number of characters contained in a syllable. Looking for other researcher's indicators of the number of characters per syllable, we found three pairs of $L_{SY}$ values for English and Spanish. First: in her PhD thesis (Barrio Cantalejo, 2008) explains how Szigriszt used Eaton's dictionary (Eaton, 1940) to estimate values $C_{SY} = 1.69$ and $C_{SY} = 2.67$ for English and Spanish respectively. Second: in their study (Trauzettel-Klosinski, Susanne; Dietz, Klaus; Group, the IReST Study, August 2012) measured the number of words, syllables and characters for 17 languages. They obtained values of $C_{SY} = 3.15$ and $C_{SY} = 1.9$ for English and Spanish. Third: (Gualda Gil, 2013) compared the density of information conveyed by English and Spanish texts. As part of his analysis, he reports values of $C_{SY} = 3.57$ and $C_{SY} = 2.94$ for English and Spanish respectively. Observing the lack of coherence among these values, we did our own count of syllables over a sample texts and calculated values of $C_{SY}$. Our results were within a 5% difference from those reported by Gualda Gil and therefore, we used the values he reported into Equation (13).

## 2.8 Message selection and groups

This study is based on written texts from historic famous speeches available in the web. The texts were originally written in different languages including English, Spanish, Portuguese, French, Italian, German, Japanese, Arabic, Russian, Chinese and Swedish. Since the analysis was done for English and Spanish, many of the texts used are translations from the original versions. Most texts are from politicians, human rights defenders and Literature Nobel laureates. We selected speeches to keep our library texts, as close as possible to the genuine writing ability of the author. Some other writing genres as the novel, have expressions from personages who distort the writing capabilities of the author. Thus, we have restricted the texts to analyze, to speeches.

We created groups of speeches and novel segments for English and Spanish: one group of texts with undoubtedly good language users, those who received Nobel Prizes for Literature, and another group by authors for which we have no special reason to assume an out-of-average use of the language. Texts classified as written by a Nobel laureate are all in their original language.

[1] In psycholinguistics a language is considered *orthographically deep* when there is little consistency between its written and spoken form. Some deep languages are Hebrew, English and French. Serbo-Croatian and Italian are examples of shallow languages (van den Bosch, Content, Daelemans, & de Gelder, 1994).

[2] Natural alphabetic languages are those whose writing consists of words build up with syllables represented by characters of an alphabet. Most known alphabets are the Latin, Cyrillic, Greek and Arabic. Natural syllabic languages represent phonograms by a single symbol or morpheme, therefore a set of letters (graphemes) to represent single sounds is not needed. Natural syllabic languages include Chinese and Japanese.

Some speeches written by Literature Nobel laureates, and translated from their original languages to English or Spanish are included in the graphs of Figures 1 to 8. These texts are clearly signaled by different markers, and are used only to obtain some sense of the effects of translations over texts authored by Nobel laureates and then subject to a translation process, but these translated Nobel laureate texts are not considered in anyone of the computation or our comparison.

To compute word frequencies, we considered punctuation signs as words. A detailed explanation about especial symbol considerations can be found in (Febres, Jaffé, & Gershenson, 2015). Text libraries, computations and results registering were administered by *MoNet*, a complex-system analysis framework we have developed to elaborate and combine results from the network of experiments which constitute this and previous works.

# 3 Results

## 3.1 Diversity for Literature Nobel laureates and for general writers

Figure 1 shows how diversity varies with the message length. The parameters of models expressed in Equations (14a) and (14b), which express Heaps' Law (Sano, Takayasu, & Takayasu, 2012) ), were adjusted by (Febres, Jaffé, & Gershenson, 2015) to minimize the summation of squared errors between the data and each model. The result is represented by the black lines in each graph of Figure 1.

$$English: \quad D_m = 3.766 \cdot L^{0.67} \, , \tag{14a}$$

$$Spanish: \quad D_m = 2.3 \cdot L^{0.75} \; . \tag{14b}$$

Notice that messages coming from Nobel laureate writers tend to appear in the higher-diversity side of the regression line defined by Equations (14a) and (14b), suggesting the possibility of grading quality of writing around diversity values.

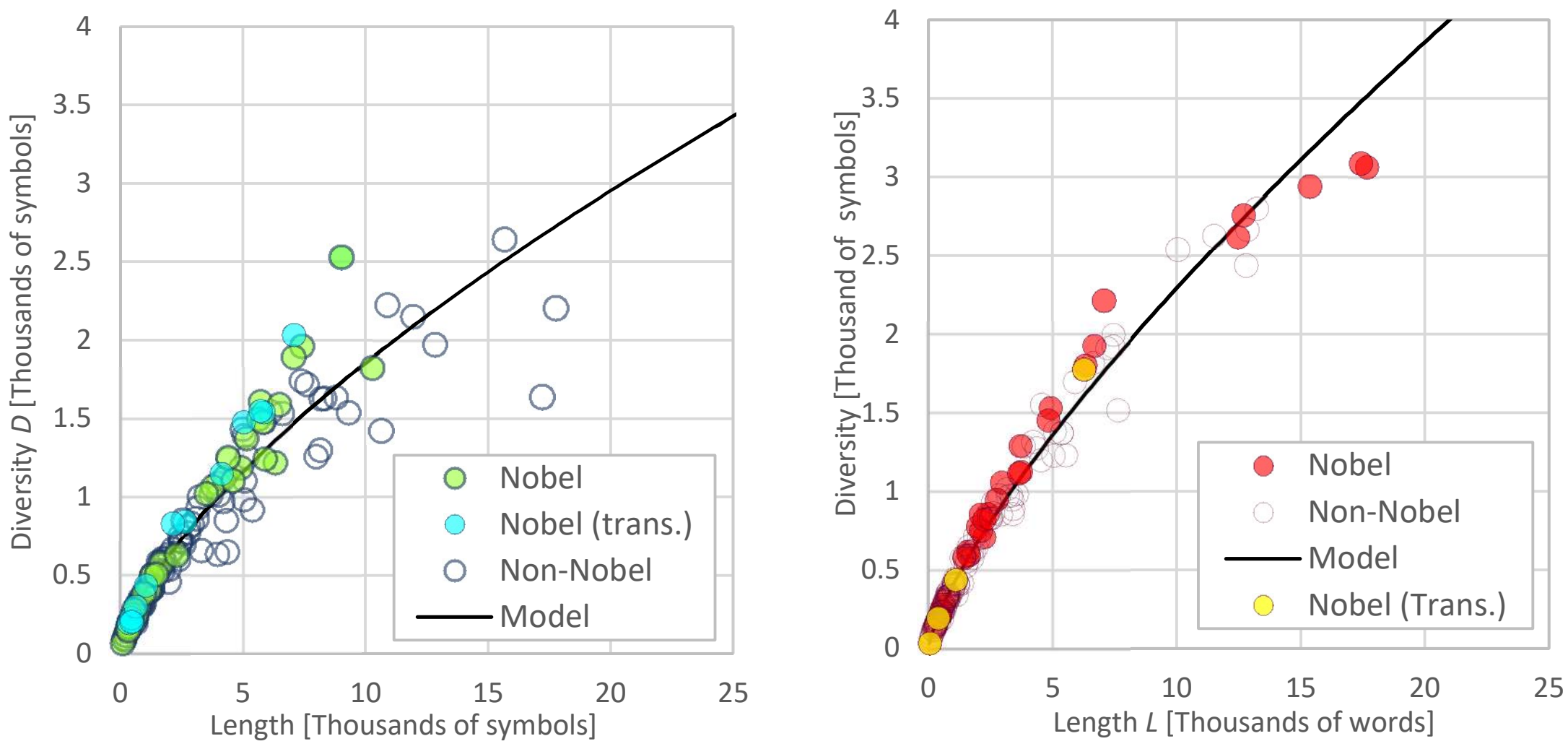

**Figure 1:** Diversity $D$ as a function of message length $L$ for messages expressed in English (left) and Spanish (right) by non-Nobel and Literature Nobel laureates. Texts authored by Literature Nobel laureates are highlighted with filled markers.

The differences between messages diversity $D$ and the diversity model expressed in Equations (14a) and (14b) was evaluated statistically for English and Spanish. Comparisons of these differences for non-Nobel laureates and Literature Nobel laureates, are shown in Table 1. The upper sector of Table 1 shows a comparison of diversities for texts written by Nobel and non-Nobel laureates. While the row for writers shows negative values for relative diversity deviations, the counterpart row for Nobel laureates, show positive values, confirming the tendency of Nobel laureate writers to use a richer vocabulary for both, English and Spanish.

| Relative specific diversity *drel* | | | | |
|---|---|---|---|---|
| | | n | *drel* *average* | *drel* *std.dev.* |
| **English:** | **Nobel laureates** | 37 | 0.02690 | 0.152 |
| **English:** | **Non Nobel** | 101 | -0.05741 | 0.133 |
| **Spanish:** | **Nobel laureates** | 19 | 0.07296 | 0.097 |
| **Spanish:** | **Non Nobel** | 117 | -0.02339 | 0.085 |
| **t-test** | | **n1 - n2** | **p-value** | |
| **English:** | **Nobel - Non Nobel** | 37 - 101 | **0.00186** | |
| **Spanish:** | **Nobel - Non Nobel** | 19 - 117 | **0.00001** | |
| **Nobel:** | **English - Spanish** | 37 - 19 | 0.23604 | |
| **Non Nobel:** | **English - Spanish** | 101 - 117 | **0.02340** | |

**Table 1:** Comparing the relative specific diversity $d_{rel}$ for English and Spanish messages by non-Nobel and Literature Nobel laureates. Upper section of the table shows the average relative specific diversity and its standard deviation. Lower section shows the p-values for Student t-tests applied to different group combinations.

When comparing English and Spanish for categories non-Nobel and Nobel laureate, the p-values are very low (especially for Spanish), meaning that the null-hypothesis should be rejected. This indicates that in English and Spanish there is a relevant difference between the relative deviation of the specific diversity $d_{rel}$ , in the texts written by Nobel and non-Nobel writers.

On the other hand, p-values for comparisons between non-Nobel and Nobel laureates indicate values sufficiently low to reject the null-hypothesis for English and Spanish. According to this, the relative deviations of the specific diversity $d_{rel}$, behave differently and offer information useful to recognize whether or not a text was written by a Literature Nobel laureate. Results show that Spanish Nobel laureates differ from other Spanish writers more than the English colleagues. Non-Nobel laureates did not differ between Spanish and English writers.

## 3.2 Entropy for Literature Nobel laureates and general writers

Figure 2 shows entropy h values for speeches expressed in natural languages versus the specific diversity d. Blue rhomboidal dots represent English messages and red circular ones represent Spanish messages. Entropy must drop down to zero when diversity decreases to zero. It also tends to a maximum value of 1 as specific diversity approaches 1. Therefore the entropy of any message can be modelled as a function of its specific diversity (Febres, Jaffé, & Gershenson, 2015), according to

$$h = \left(\frac{D}{L}\right)^{(\alpha-2)/(\alpha-1)} = d^{(\alpha-2)/(\alpha-1)} , \qquad (15)$$

where $\alpha$ is a real number. Expressions (16a) and (16b) were obtained after adjusting parameter $\alpha$ to fit experimental data.

$$Engli \;:\quad h = d^{0.1523}\,, \tag{16a}$$

$$Spanish: \quad h = d^{0.1763}\;. \tag{16b}$$

Figure 2 also differentiates between writers and Nobel laureates. Color filled dots represent speeches written by Literature Nobel laureate. Messages originated by other writers are represented by empty dots.

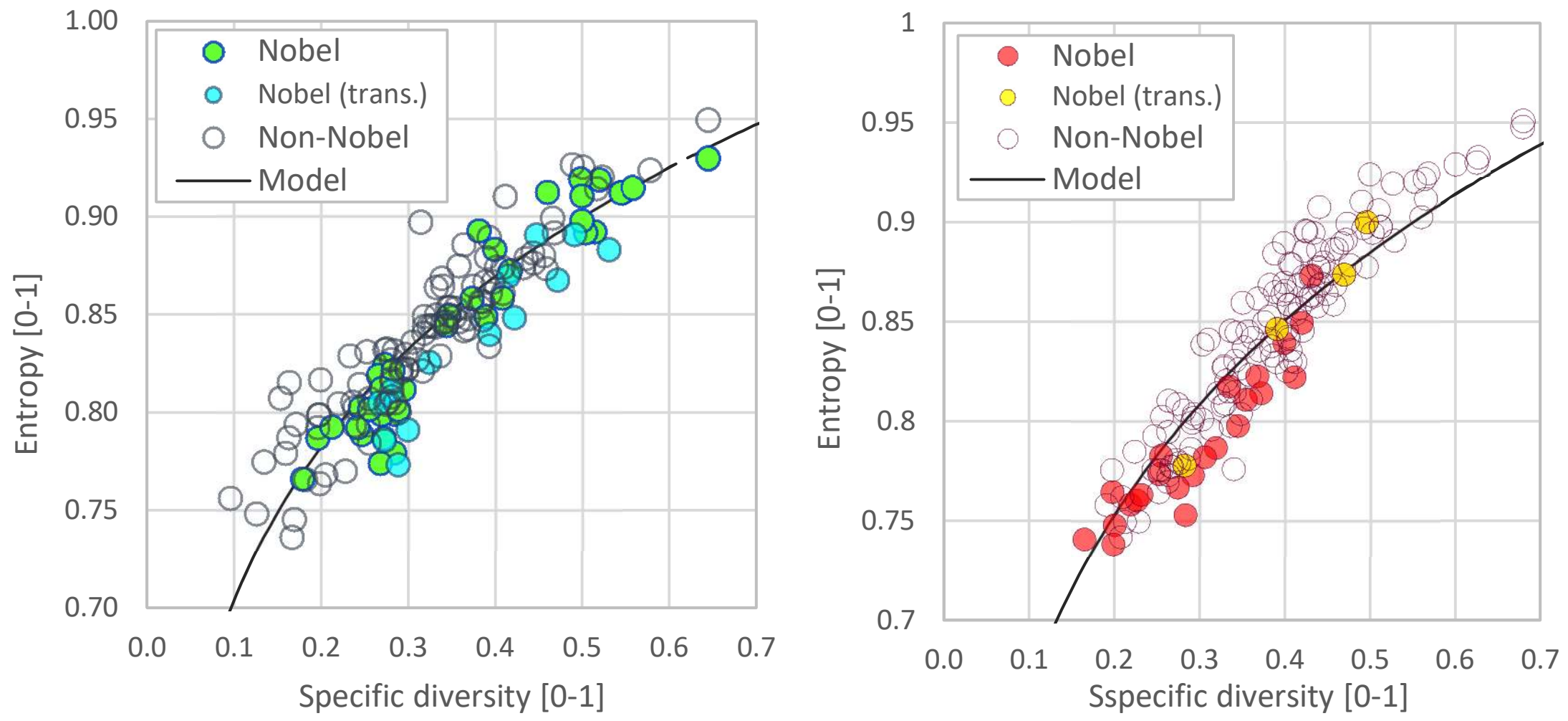

**Figure 2:** Entropy $h$ vs. specific diversity $d$ for messages expressed English (left) and Spanish (right) by non-Nobel and Literature Nobel laureates. Texts authored by Literature Nobel laureates are highlighted with filled markers.

| **Relative entropy *hrel*** | | | | |
|---|---|---|---|---|
| | | **n** | ***hrel* *average*** | ***hrel* *std.dev.*** |
| **English:** | **Nobel laureates** | 37 | -0.00567 | 0.0192 |
| **English:** | **Non Nobel** | 101 | 0.00318 | 0.0184 |
| **Spanish:** | **Nobel laureates** | 19 | -0.01168 | 0.0192 |
| **Spanish:** | **Non Nobel** | 117 | 0.00579 | 0.0183 |
| **t-test** | | **n1 - n2** | **p-value** | |
| **English:** | **Nobel - Non Nobel** | 37 - 101 | **0.00659** | |
| **Spanish:** | **Nobel - Non Nobel** | 19 - 117 | **0.00005** | |
| **Nobel:** | **English - Spanish** | 37 - 19 | 0.2067 | |
| **Non Nobel:** | **English - Spanish** | 101 - 117 | 0.2852 | |

**Table 2:** Comparing the relative entropy $h_{rel}$ for English and Spanish messages by non-Nobel and Literature Nobel laureates. Upper section of the table shows the average relative entropy and its standard deviation. Lower section shows the p-values for Student t-tests applied to different group combinations.

It is visually noticeable that dots representing texts from Nobel laureates tend to lie at a lower entropy level than that indicated by the lines representing models (16a) and (16b). Nobel laureate texts show less entropy than the average for non-Nobel laureates in both Spanish and English. The difference between the two categories was analyzed statistically and results are shown in Table 2. Relative entropies $h_{rel}$ for non-Nobel writers and Nobel writers show opposite signed values. This difference in the distribution of relative entropy for writers and Nobel laureates is confirmed by the Student t-test; p-values printed in bold numbers are very low and therefore the hypothesis is rejected for English and Spanish.

### 3.3. Zipf's deviation $J_{1,D}$ for ranked distribution

Profile of symbol frequency distributions were inspected in two ways: first by a qualitative analysis of their shapes, and second by characterizing each profile with its area deviation J with respect to a Zipf distributed profile. A sample of symbol frequency distribution profiles for the considered languages is represented in Figure 3. Each sequence of markers belongs to a message and each marker corresponds to a word or symbol within the message. The size of the sample included in Figure 3 is limited to avoid excessive overlapping of markers which would keep from appreciating the shape of each profile. No important differences are observed among messages profiles expressed in the same language, however.

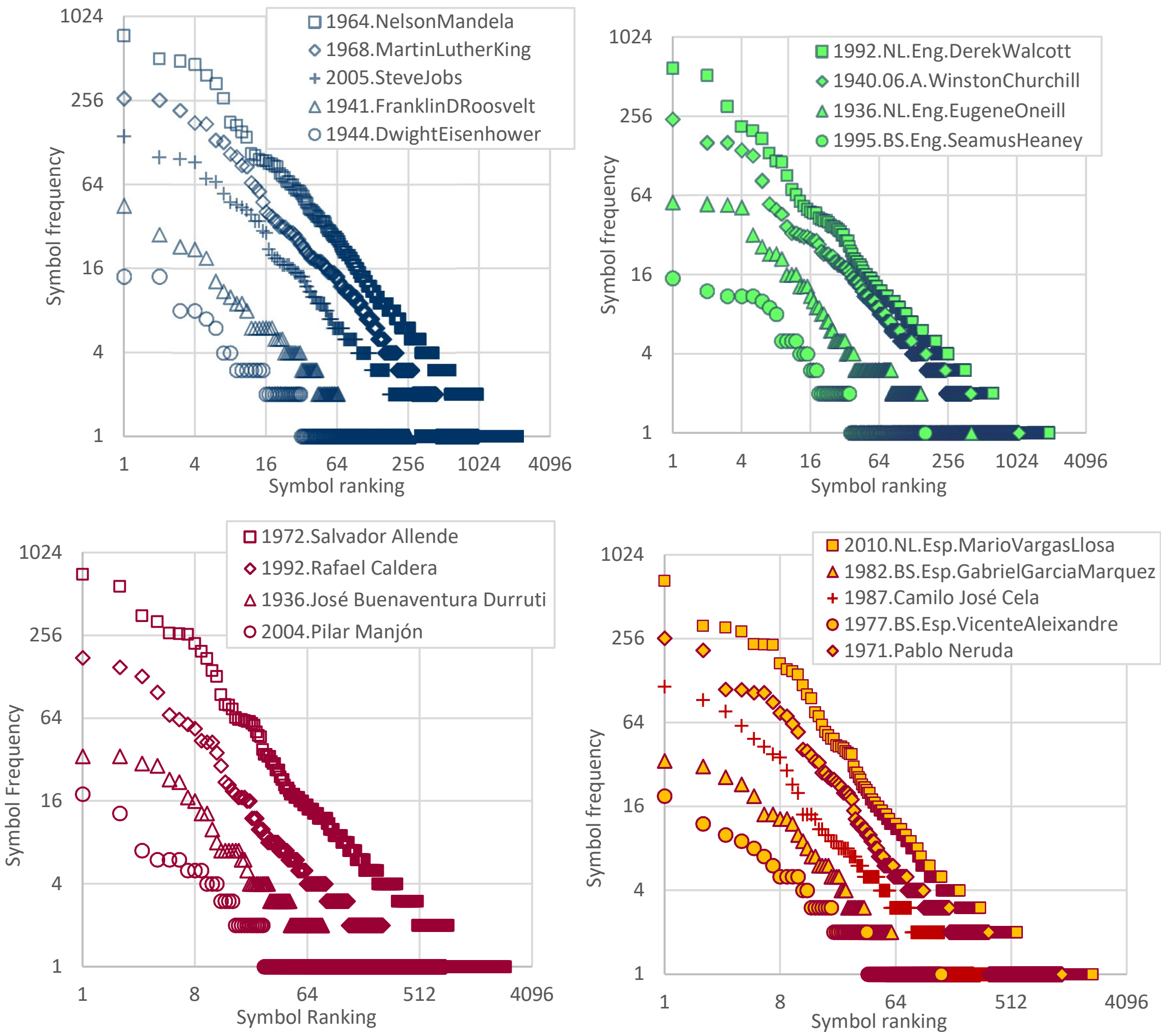

**Figure 3:** Ranked symbol frequency distribution profiles. Sample of three profiles for each category. Upper row shows English message profiles and lower row Spanish message profiles. Left column graphs show the profiles for writer originated texts and right column Nobel laureate texts.

Zipf's deviations $J_{1,D}$ for messages written by writers and Literature Nobel laureates are illustrated in Figure 4. Messages written by Literature Nobel laureates, exhibit lower values of Zipf's deviations $J_{1,D}$ in comparison to Zipf's deviation of texts coming from non-laureate writers. To measure this difference, we computed Zipf's deviations $J_{1,D}$ for different groups of data: languages and writers class. Table 3 summarizes these results.

Spanish texts from Nobel-laureates show different Zipf's deviations when compared with texts from non-Nobel writers. For English texts, this difference is more subtle than the difference when the language is Spanish. Comparing the non-Nobel with Nobel writers, the p-value for Spanish is less than 0.00003, low enough to reject the null hypothesis, meaning that for Spanish the deviation of the Zipf's distribution is different for the two writer categories considered. For English the p-value of 0.00396, is also sufficiently low to reject the null hypothesis between these two categories. In fact, average values $J_{1,D}$ for English-non-Nobel writers (0.03232) and English Nobel (-0.05779) are relatively far from each other. For Spanish this statistic is different; values $J_{1,D}$ for non-Nobel (-0.10382) and Spanish Nobel (-0.19167) are sensibly different.

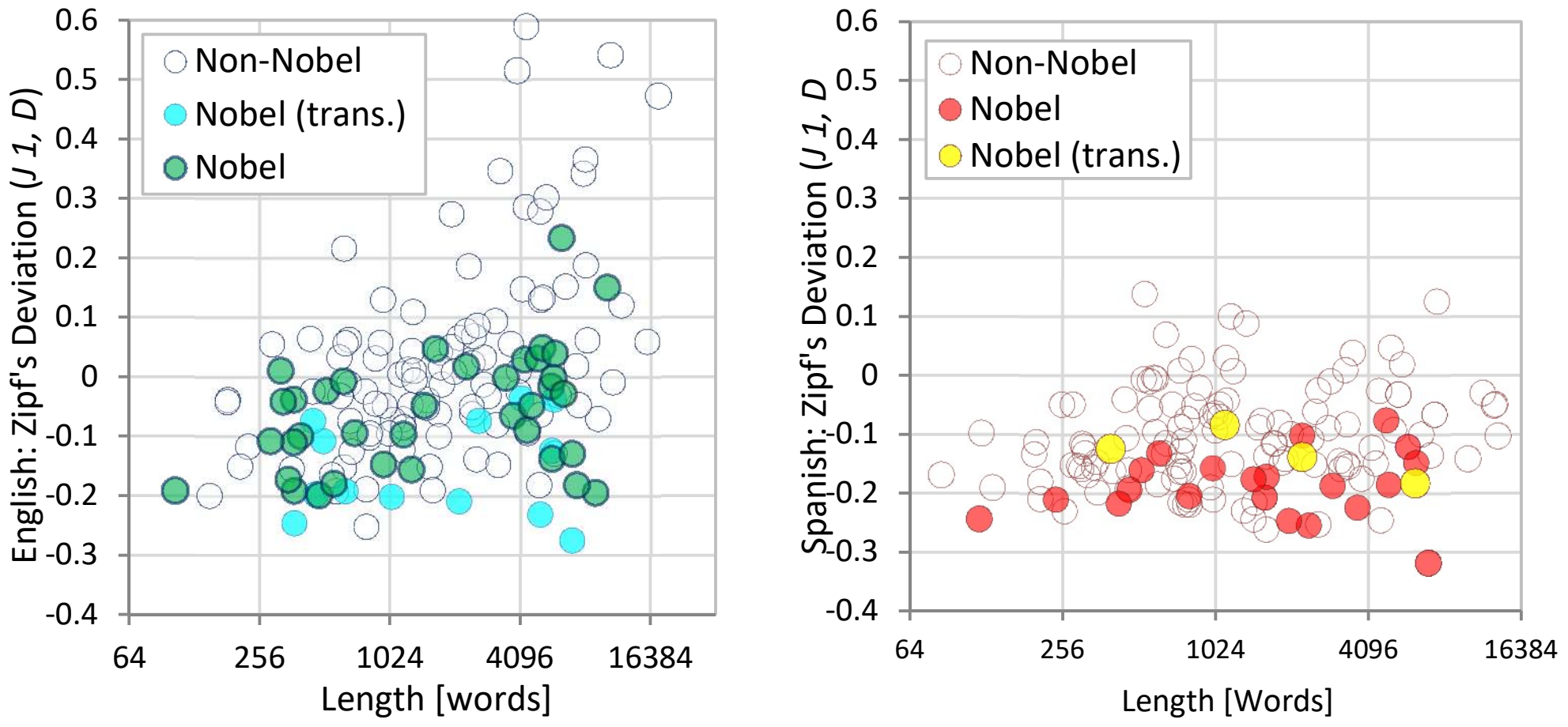

**Figure 4:** Relative Zipf's deviation $J_{1,D}$ vs. message length $L$ for messages expressed in English (left) and Spanish (right) by non-Nobel and Literature Nobel laureates. Texts authored by Literature Nobel laureates are highlighted with filled markers.

| Relative Zipf's deviation $J_{1,D}$ | | | | |
|---|---|---|---|---|
| | | n | $J_{1,D}$ average | $J_{1,D}$ std.dev. |
| English: | Nobel laureates | 37 | -0.05779 | 0.0994 |
| English: | Non Nobel | 101 | 0.03232 | 0.1768 |
| Spanish: | Nobel laureates | 19 | -0.19167 | 0.0561 |
| Spanish: | Non Nobel | 117 | -0.10382 | 0.0856 |
| t-test | | n1 - n2 | p-value | |
| English: | Nobel - Non Nobel | 37 - 101 | 0.00396 | |
| Spanish: | Nobel - Non Nobel | 19 - 117 | 0.00003 | |
| Nobel: | English - Spanish | 37 - 19 | < 0.00001 | |
| Non Nobel: | English - Spanish | 101 - 117 | < 0.00001 | |

**Table 3:** Comparing the relative Zipf's deviation $J_{1,D}$ for English and Spanish messages by non-Nobel and Literature Nobel laureates. Upper section of the table shows the average relative Zipf's deviation and its standard deviation. Lower section shows the p-values for Student t-tests applied to different group combinations.

### 3.4 Writing Quality Evaluation

Not being a Literature Nobel laureate does not mean poor writing capabilities. But winning a Literature Nobel Prize is guarantee of being gifted for excellent writing as well as master knowledge and control over a natural language. Some measurable statistical difference should emerge from classifying writers by those who were recognized with a Nobel Prize, and those who were not.

Figures 1, 2 and 4 present a clear evidence of the tendency of speeches from Nobel laureates to differ from the average style of writing of public figures. When comparing Nobel and non-Nobel laureate messages, the average of the former group tends to show higher specific diversity $d$ and lower entropy $h$. This is interesting because the higher specific diversity of Nobel laureate texts should promote a higher entropy due to the larger scale $D$ of the language used implied by the larger vocabulary. See Equations (2) and (15) to observe how $D$ affects the resulting entropy $h$. Nonetheless, in spite of the larger vocabulary exhibited by Literature Nobel laureates in their texts, the associated entropies $h$ are lower. Thus $h_{rel}$ is a second variable to include in a writing quality evaluation scale.

Our data shows that Zipf's deviation $J_{1,D}$ is a third variable to have influence over a writing quality evaluation scale.

As some clustering is observed for the Nobel laureate class, we estimated the coordinates of the centers, and a direction vector pointing from the non-laureate class center to the Nobel laureate class center. These directions provide a sense for creating a scale that is sensitive to the quality of writing for English and Spanish. The clusters centers coordinates are:

$$English\ writers\ class: \quad (d_{rel}, h_{rel},\ J_{rel}) = (-0.05741\ ,\ \ 0.00318\ , -0.03232) \tag{17a}$$
$$English\ Nobel\ class: \quad (d_{rel}, h_{rel},\ J_{rel}) = (\ \ 0.0269\ ,\ -0.00567\ ,\ -0.05779) \tag{17b}$$
$$Spanish\ writers\ class: \quad (d_{rel}, h_{rel},\ J_{rel}) = (-0.02339\ ,\ \ 0.00579\ ,\ -0.08785) \tag{17c}$$
$$Spanish\ Nobel\ class: \quad (d_{rel}, h_{rel},\ J_{rel}) = (\ \ 0.07296\ ,\ -0.01168\ ,\ -0.19167) \tag{17d}$$

The direction vectors are:

$$English\ class\ direction\ vector: \ (d_{rel}, h_{rel},\ J_{rel}) = (0.68147\ ,\ -0.07153\ ,\ -0.72835) \tag{18a}$$
$$Spanish\ class\ direction\ vector: \ (d_{rel}, h_{rel},\ J_{rel}) = (\ 0.73241\ ,\ -0.13280\ ,\ -0.66779) \tag{18b}$$

Based on the direction vectors and the non-Nobel writer's class center coordinates, we suggest the following a Writing Quality Scale ($WQS$) which we claim is sensible to the quality of writing.

$$English: \ WQS = 5.5082\ (d_{rel} + 0.02690) - \ 0.5782\ (h_{rel} - 0.00318)\ - 5.8871\ (J_{1,D} - 0.03232) \tag{19a}$$
$$Spanish: \ WQS = \ 5.5674\ (d_{rel} + 0.02339) - \ 1.0095\ (h_{rel} - 0.00579)\ - 5.0762 (J_{1,D} - 0.10382) \tag{19b}$$

We computed $WQS$ for each text as Equations (19a) and (19b) indicate. Values of relative specific diversity $d_{rel}$, relative entropy $h_{rel}$, Zipf's deviation $J_{1,D}$ are the same included in Appendixes A, B, C, and D and graphically shown in Figure 5.

Whether or not the language of a speech or a novel, is the author's native language, may be a factor with some influence over the evaluation of the $WQS$. For the case of all statistics in this study, a speech or novel is considered to be authored by a Nobel Prize winner only in the version the text is presented in the author's native language. That choice assumes that the difference, which can be subtle, between the style of writing of a Nobel laureate and a non-Nobel writer, could vanish in the process of translation. The selected criterion for considering a text written by a Literature Nobel laureate, evades possible effects of the translation, when it is performed, over the statistics presented and also over the models we call $WQS$. However, in Figure 5, those texts originally written by Nobel laureates and thereafter translated into English or Spanish, are included in the left graphs together with the texts authored by Nobel laureate writers. In the left graphs of Figure 5, the bubbles representing texts by Nobel laureates, are bordered with a thick dark ring, while translated texts have a thin border.

Representing Nobel laureate texts and translated texts from original Nobel laureate texts, allows for visually appraise the impact over the $WQS$ value of texts translations. The left graphs in Figure 5 suggest that there is no important difference among the $WQS$ values for original and translated Nobel laureate texts. Comparing left graphs with right graphs, there is a noticeable tendency for the Spanish texts written by Literature Nobel laureates, to cluster around the point signaled by Expression (17d). English texts do not show as much clustering as Spanish texts do, which is consistent with the p-values of the Student t-test shown in Table 3.

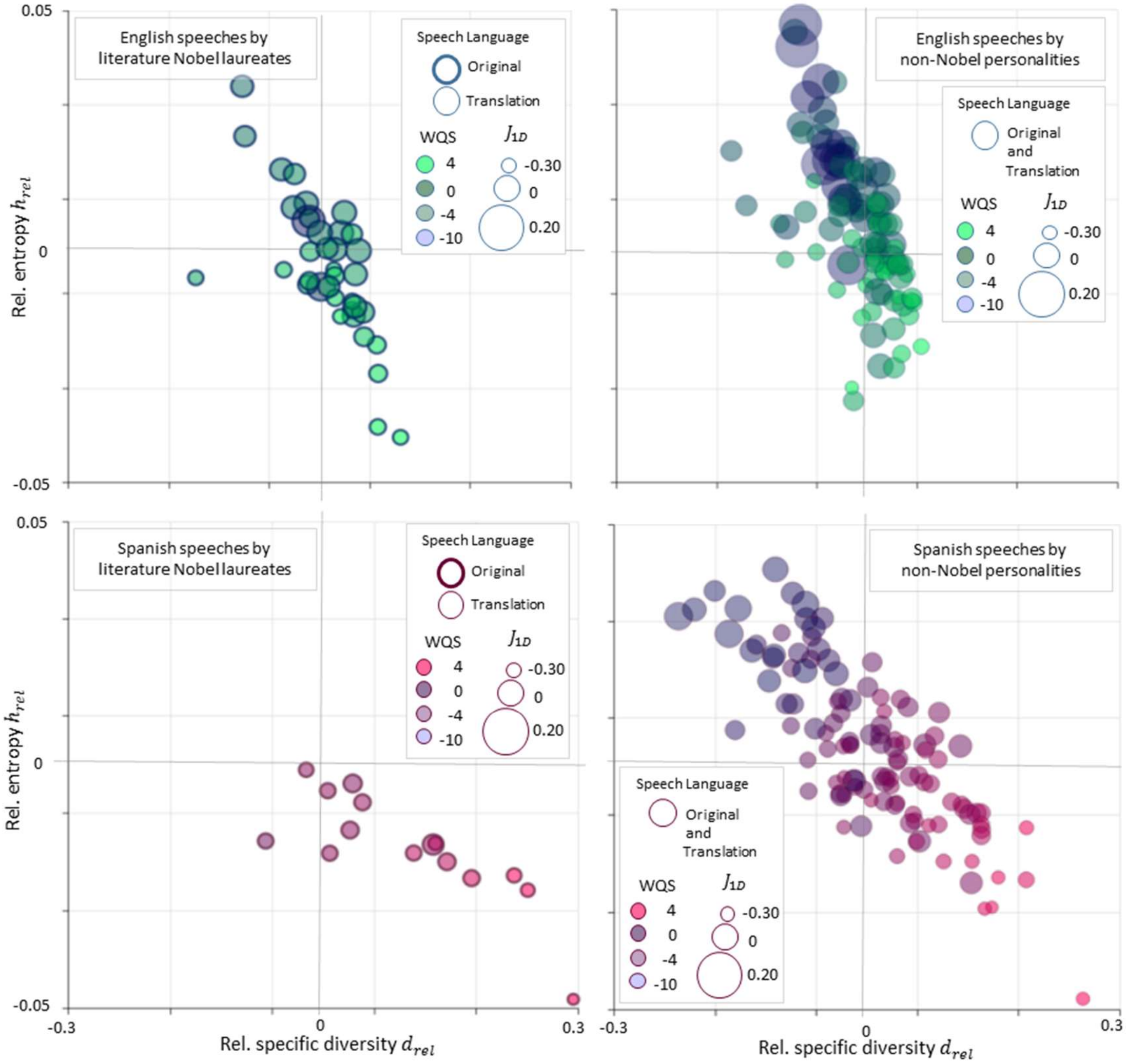

**Figure 5:** Writing quality evaluation for English (top) and Spanish (bottom) texts. Left graphs correspond to Literature Nobel laureate texts and right graphs to non-Nobel texts. Horizontal axes represent relative specific diversity $d_{rel}$. Vertical axes represent relative entropy $h_{rel}$. The Zipf's deviation $J_{1,D}$ is represented as proportional to the radius of the bubbles. The writing quality scale $WQS$ value is represented by the bubble color.

### 3.5 Writing Quality Scales and Readability indexes

The Writing Quality Scales ($WQS$) developed in section 3.4 were compared with the readability indexes from Flesch ($RES$) and Szigriszt ($IPSZ$). Figure 6 shows graphs of readability indexes versus the $WQS$ obtained for each text in the library. In the graphs, each dot represents a text. To enable the graphs to visually show the difference between text categories, filled dots correspond to texts written by Literature Nobel laureates and empty dots show texts by non-Nobel writers. For Spanish, there is a higher density of dots representing texts by Nobel laureates towards the high $WQS$ region, placed to the right of horizontal axis. For English, texts written by Nobel laureates and non-Nobel do not show any important difference in their dispersion over the space of any axis.

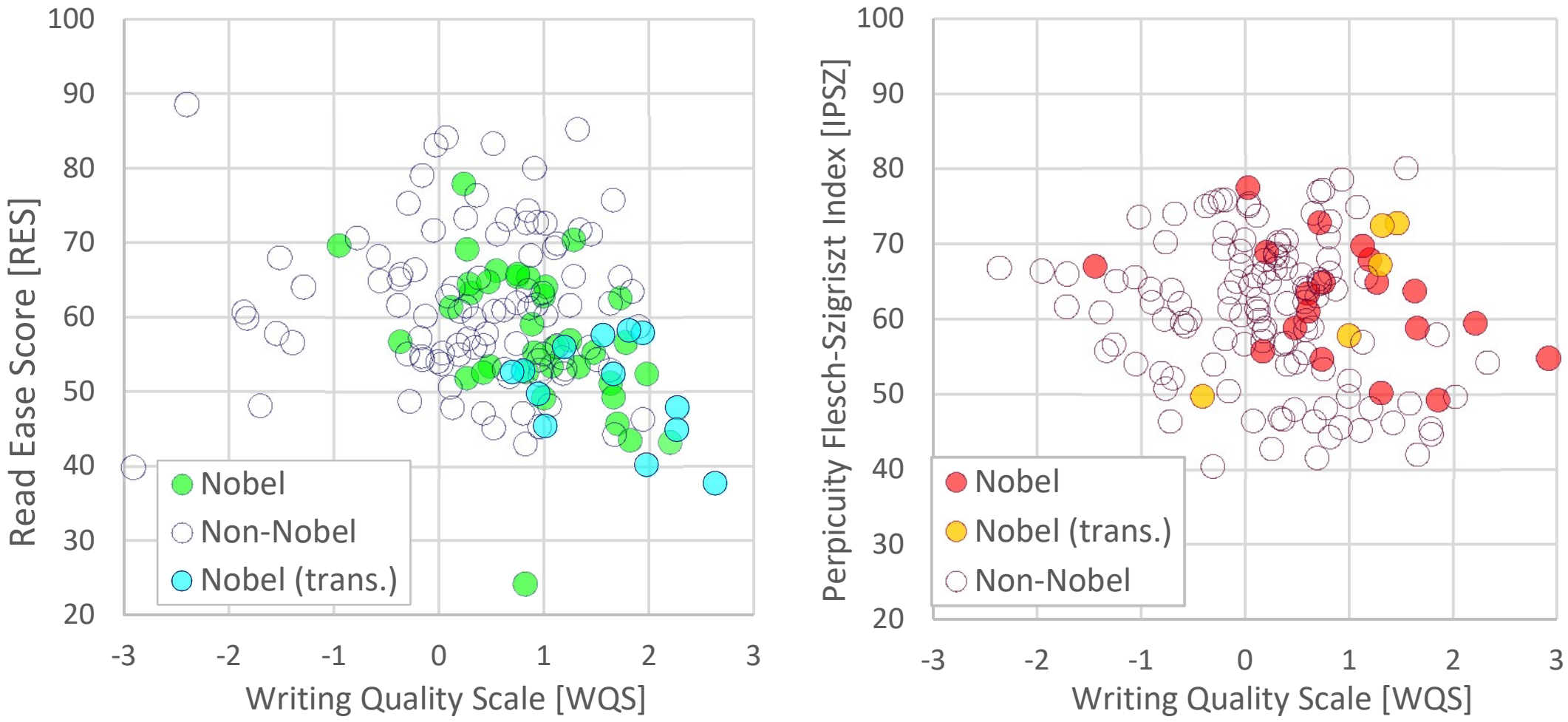

**Figure 6:** Text readability vs. Writing Quality Scale $QWS$ for English texts (left) and Spanish texts (right). English readability is measured as Flesch $RES$ (Reading Ease Score). Spanish readability is measured as Szigriszt $IPSZ$ (Perspicuity Index). Filled dots (green for English texts and orange for Spanish texts) correspond to texts written by Literature Nobel laureates. Empty dots correspond to non-Nobel texts.

| Writing Quality Scale and Readabilty | | | | | | | |
|---|---|---|---|---|---|---|---|
| | | n | *WQS average* | *WQS std.dev.* | *Readability* average [0-100]* | *Readability std.dev.* | *Correlation WQS-Readty.* |
| **English:** | **Nobel and non-Nobel** | 138 | 0.45 | 1.13 | 59.74 | 10.45 | -0.36 |
| | **Nobel laureates** | 37 | 1.00 | 0.82 | 56.30 | 9.71 | -0.40 |
| | **Non -Nobel** | 101 | 0.25 | 1.16 | 61.00 | 10.47 | -0.31 |
| **Spanish:** | **Nobel and non-Nobel** | 136 | 0.11 | 0.98 | 62.19 | 10.45 | -0.27 |
| | **Nobel laureates** | 19 | 0.22 | 1.28 | 70.41 | 10.26 | -0.70 |
| | **Non -Nobel** | 117 | 0.09 | 0.92 | 60.85 | 9.89 | -0.21 |
| **t-test** | | **n1 - n2** | **p-value** | **t-test** | | **n1 - n2** | **p-value** |
| **English WQS:** | **Nobel - Non Nobel** | 37 - 101 | 0.000 | **Spanish WQS:** | **Nobel-Non Nobel** | 19 - 117 | 0.582 |
| **English RES:** | **Nobel - Non Nobel** | 37 - 101 | 0.019 | **Spanish IPSZ:** | **Nobel-Non Nobel** | 19 - 117 | 0.000 |

* Readability is meassured as *REF* for English and *IPSZ* for Spanish.

**Table 4:** Comparing the Writing Quality Scale $WQS$ for English and Spanish messages by non-Nobel and Literature Nobel laureates. Upper section of the table shows the average Writing Quality Scale and its standard deviation. Lower section shows the p-values for Student t-tests applied to different group combinations.

Numerical comparisons between these different texts evaluations, are included in Table 4, confirming the visual appreciation mentioned above. Even though small, there is a difference between the averages of the distributions of Spanish readability indexes $IPSZ$ for texts authored by Nobel writers and non-Nobel writers. At the same time, the small p-value obtained from Student-t tests for these distributions, indicates they are different, and that Literature Nobel laureates tend to produce more readable texts than others writers. The Student-t test performed between the distributions of English readability indexes $RES$ for Nobel and non-Nobel texts, resulted in a high p-value indicating that there is not any important difference between these distributions of the readability index.

For English and Spanish texts, values of $WQS$ for Nobel laureates showed higher when compared with values for texts coming from non-Nobel writers. Probable differences between distributions of the $WQS$ of texts written in English and Spanish, were evaluated by Student t-tests. The results of these tests, included in Table 4, indicate that for English, Nobel and non-Nobel $WQS$ values are likely to come from different distributions, while for Spanish these distributions are definitively different.

## 3.6 Writing style change in time

Even though our main objective is to compare differences in writing style, we take advantage of the data fed, in order to investigate the change that the average sentence length, as measured in words, and the writing quality scale $WQS$, may have had over the last couple of centuries.

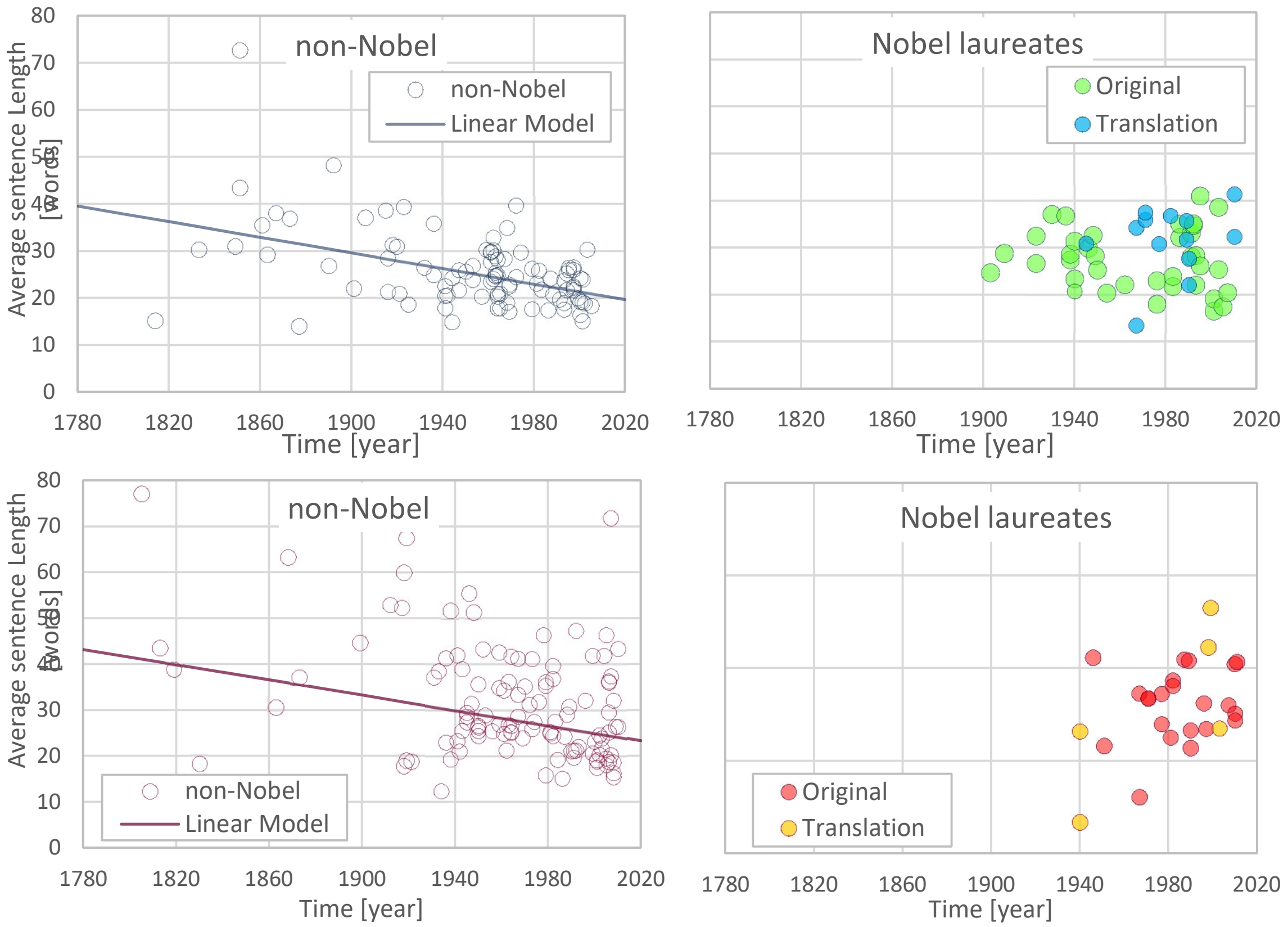

**Figure 7:** Average sentence length [words] vs. year when the speech was written for English (top row) and Spanish (bottom row). Left graphs show texts written by non-Nobel writers. A continuous line represent an error minimum summation model. Graphs on the right show Nobel laureate speeches. Original and translated texts are represented with different markers.

Figure 7 presents the average number of word per sentence for each speech of our texts library. Figure 8 shows the Writing Quality Scale $WQS$ values. In both Figures, four graphs are presented for the combinations of speeches expressed in English and Spanish, and authored by non-Nobel writers and Literature Nobel laureate speeches. The speeches authored by Nobel laureates, but resulting from translation from another language, are included in the graph dedicated to Nobel laureates. They are distinguished with a different marker from the one used for Nobel laureate texts expressed in their original language.

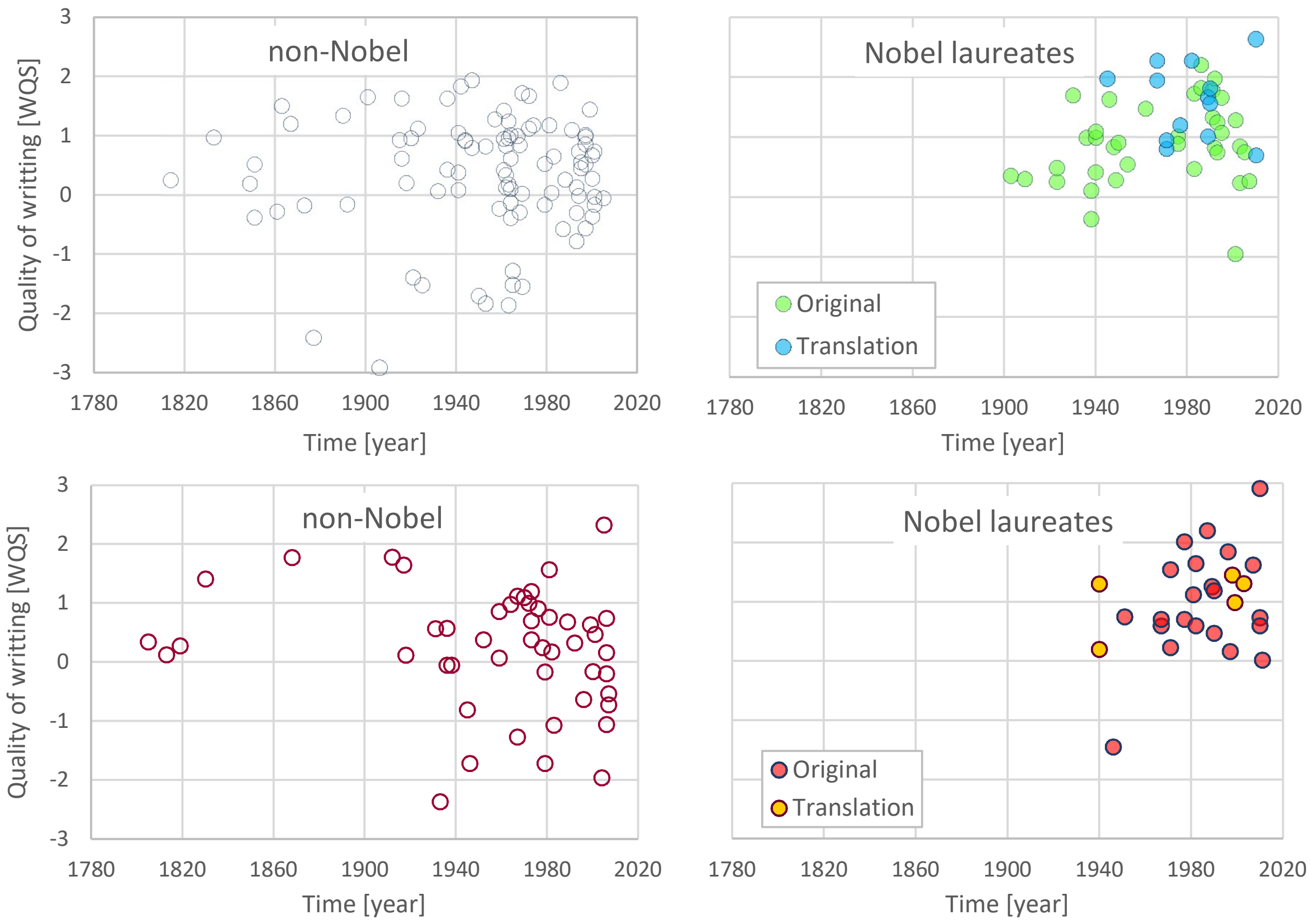

**Figure 8:** Writing Quality Scale $WQS$ vs. year when the speech was written for English (top row), and Spanish (bottom row). Graphs on the left show texts written by non-Nobel writers. Graphs on the right show Nobel laureate speeches. Original and translated texts are represented with different markers.

None of the graphs of Figure 8 exhibit any important tendency of the $WQS$ over time. But the average sentence length for non-Nobel writers graphed in Figure 7, does show a clear tendency to diminish over time for both, English and Spanish. Thus, we included a line to show the resulting regression from the minimization of the summation of the errors. Those line equations are $L = 187.1 - 0.0829\, y$ for English, and $L = 189.8 - 0.0824\, y$ for Spanish where $L$ is the average lenth of sentences and $y$ is the year.

# 4 Discussions

## 4.1 Diversity and Entropy

In general, Literature Nobel laureates exhibit a richer vocabulary in their speeches when compared with other writers. Clearly, a necessary condition to win a Nobel Prize is the knowledge of an extended lexicon and the wisdom to use it appropriately and with a well-organized style. The higher diversity of words in exhibited by most texts from Nobel laureate shown in Figure 1, is thus, an expected result. Interestingly, Nobel laureates somehow handle this higher word diversity in such a way, that they produce texts with considerably lower entropy than the expected entropy value, at the corresponding specific diversity. Therefore, the lower entropy values exhibited by Nobel laureate's texts, must obey to the word's frequency distribution they use, which overcomes the natural effect of the larger diversity of words present in their texts.

### 4.2 Symbol Frequency Distribution Profile

The difference between the Zipf's deviations $J_{1,D}$ for the two types of writers originating the texts, is small. However, the relatively small p-values indicate that Zipf's deviations $J_{1,D}$ express some of the differences between texts originated by Nobel and non-Nobel writers, and therefore, the inclusion of the Zipf's deviations $J_{1,D}$ as a writing quality sensitive factor, is justified.

### 4.3 Writing Quality Scale versus Readability Index

Readability indexes are intended to classify the ease with which a text can be read and understood. They are not directly associated to quality or style of writing. In fact, evaluating quality and style of writing is a highly subjective matter, difficult to submit to a quantifying procedure. It is a subtle and elusive task. However, good writing structure and style must include readability as an important characteristic of the resulting text. The measures of entropy explored here add information about more general aspects of writing quality.

Another factor influencing the readability indexes is the complexity of the idea being deployed with the text. A complex idea, probably, cannot be explained with the same high readability index of a simple idea. Thus the question is: What readability index can reach a writer when he or she writes a text to convey some complex idea? There is no obvious answer, among other reasons, because the complexity of the idea itself, is a subjective factor. But good writers should tend to produce more readable texts — with higher readability indexes: $RES$ for English and $IPSZ$ for Spanish— than those less talented for this activity. In fact, Figure 6 shows that for Spanish there is higher density of texts authored by Nobel laureates over the higher readability region, indicating that Spanish Nobel laureates tend to produce high readability texts. For English, we did not detect any important difference between the readability of the texts produced by Nobel and non-Nobel laureates.

Figure 5 illustrates how most of the texts in Spanish with high values of $WQS$, those which are reddish, lie in the lower right quadrant. This quadrant represents texts with lower relative entropy and higher specific diversity; both tendencies formerly associated with the style of writing of Literature Nobel laureates. A similar orientation of the $WQS$ value, is observed for English written texts, even though it is not as notorious as it was for Spanish texts. This confirms that the $WQS$ captures some of the properties associated quality and style of writing. Especially for Spanish writing.

### 4.4 Tendencies of the writing style

The change of the average sentence length estimated from the regression model shown in Figure 7, is a reduction of 8.29 and 8.24 words per century for sentences written in English and Spanish respectively; interestingly two values that are, in the practical sense, equal.

According to previous results by Sherman (DuBay, 2004), the sentence length experienced a change of 22 words (from 45 to 23), in a time span of 293 years, from the times of Queen Elizabeth I (around the year 1600) to Sherman's times (around the year 1893). These numbers and dates result in a calculated decrease of 7.5 words per century for English; a figure consistent with our estimates, which validates the comprehensiveness of the data we used.

Independently of the results from Sherman's works, the reduction of the length of sentences observed in Figure 7, seems to be a sustained tendency for common writers. Perhaps the increasing need to produce more effective texts, leaving less space for words dedicated to embellish the texts, is partially responsible for the reduction of the number of words. The evolution of the natural languages may also contribute, by the acceptance of new words, to the representation of concepts and ideas in a more compact fashion. Nevertheless, the decrease of the number of words in a typical sentence, is probably approaching a lower bound, since a certain number of words is needed to express precise and elaborated ideas.

The Nobel Prizes are awarded since 1901. The history records we have to evaluate the evolution of styles on Literature Nobel laureates, are shorter than the sample of speeches we have available for non-laureate writers. Yet, the average sentence length for Nobel laureate writers does not show any important tendency to change over time. This suggests that good style of writing is not necessarily aligned with the concept of readability. There is no obvious increase or decrease of the values of $WQS$ in the graphs included in Figure 8. This suggests there is no direct incidence of the sentence length over the value of the Writing Quality Scale $WQS$.

# 5 Conclusions

Our analysis showed that some properties of texts written in English and Spanish, such as entropy, symbol diversity, and frequency distribution profiles, relate to aspects of what is considered by professionals as "good writing" in natural languages. In general, our method showed to work better for Spanish than for English language. Texts written in Spanish by Nobel laureates and non-Nobel, are easier to segregate than their counterpart in English. The visual assessment of graphs as well as the statistical evaluations, confirm this statement. However, even for the English language, the method is capable to classify a text according to its writing quality as compared with a text representative of those written by a Literature Nobel laureate. This is encouraging because it suggests the feasibility of using quantitative measures to characterize certain aspects related to the quality of writings.

This opens the door to eventually develop tools for automatic text evaluations. The fact that quality was related to higher specific diversity and less entropy, suggests that skillful writing involves incorporation of order into the text. The precise nature of this additional order is still unknown, but our method serves to detect its presence.

The results found so far are to be taken as insights of a preliminary exploration of the complexity of texts. Certainly, further studies applying these methods to a larger set of texts and extending the methods to other writing genres may lead to further refinements that may make $WQS$ a useful tool for evaluation of writing capabilities. We believe, however, that feasibility of automated quantitative evaluation of writing quality is getting closer.

# Acknowledgements

We should like to thank an anonymous referee for useful comments which let us to enrich our work.

# Appendix A

**English texts: Literature Non-Nobel laureates. Readability and Writing Quality Scale comparison**

*G = Genre* [S=Speech: N Novel segment/Story] | ***h*** Entropy [0-1] | ***J1,D*** Zipf's deviation

*L = Language* [ E= English : Translation to E] | ***drel*** rel. specific diversity [0-1] | ***RES*** Flesch Read. Easy Score

*d = specific diversity [0-1]* | ***hrel*** rel. entropy [0-1] | ***WQS*** Writing Qualirty Scale

| *Index* | *Text Name* | *G* | *L* | *d* | *h* | *drel* | *hrel* | *J1,D* | *RES* | *WSQ* |
|---|---|---|---|---|---|---|---|---|---|---|
| ET1 | 1814.NapoleonBonaparte | S | T | 0.522 | 0.920 | -0.2170 | 0.0087 | -0.0406 | 55.3122 | -0.0306 |
| ET2 | 1921.MarieCurie | S | T | 0.333 | 0.864 | -0.1462 | 0.0233 | 0.0582 | 71.7192 | -0.1625 |
| ET3 | 1941.AdolfHitler | S | T | 0.204 | 0.769 | 0.1833 | -0.0174 | -0.0076 | 76.4677 | 0.7375 |
| ET4 | 1979.MotherTeresa | S | T | 0.150 | 0.805 | -0.3591 | 0.0423 | 0.5909 | 52.0252 | -3.4579 |
| ET5 | 1994.MotherTeresa | S | T | 0.161 | 0.815 | -0.3314 | 0.0469 | 0.5166 | 68.3648 | -3.0009 |
| ET6 | 2000.PopeJohnPaulII | S | T | 0.397 | 0.875 | -0.0194 | 0.0078 | -0.0937 | 85.2700 | -0.0545 |
| E1 | 1381.JohnBall | S | O | 0.515 | 0.914 | -0.1684 | 0.0049 | -0.1156 | 59.9047 | 0.6114 |
| E2 | 1588.QueenElizabethI | S | O | 0.435 | 0.879 | -0.1844 | -0.0027 | -0.1528 | 74.4296 | 0.8685 |
| E3 | 1601.Hamlet | S | O | 0.647 | 0.950 | -0.0899 | 0.0139 | -0.1987 | 73.3751 | 1.3180 |
| E4 | 1601.QueenElizabethI | S | O | 0.340 | 0.865 | -0.0646 | 0.0209 | 0.0056 | 53.0996 | 0.3335 |
| E5 | 1606.LancelotAndrewes | S | O | 0.166 | 0.737 | -0.0893 | -0.0326 | -0.0690 | 55.0698 | 0.8442 |
| E6 | 1833.ThomasBabington | S | O | 0.169 | 0.746 | 0.1025 | -0.0253 | 0.0602 | 65.1686 | 0.2549 |
| E7 | 1849.LucretiaMott | S | O | 0.227 | 0.771 | 0.1658 | -0.0256 | -0.0457 | 83.3508 | 0.9703 |
| E8 | 1851.ErnestineLRose | S | O | 0.196 | 0.764 | 0.0391 | -0.0187 | 0.0630 | 48.8446 | 0.1891 |
| E9 | 1851.SojournerTruth | S | O | 0.413 | 0.919 | -0.1738 | 0.0455 | 0.0651 | 54.4254 | -0.3795 |
| E10 | 1861.AbrahamLincoln | S | O | 0.254 | 0.808 | 0.0572 | 0.0001 | 0.0116 | 53.1160 | 0.5157 |
| E11 | 1863.AbrahamLincoln | S | O | 0.490 | 0.927 | -0.1414 | 0.0259 | 0.0557 | 57.3512 | -0.2788 |
| E12 | 1867.ElizabethCadyStanton | S | O | 0.253 | 0.785 | 0.1940 | -0.0227 | -0.1277 | 88.6036 | 1.5001 |
| E13 | 1873.SusanBAnthony | S | O | 0.407 | 0.862 | -0.0814 | -0.0091 | -0.1505 | 71.8227 | 1.2006 |
| E14 | 1877.ChiefJoseph | S | O | 0.503 | 0.926 | -0.2445 | 0.0203 | -0.0375 | 54.3828 | -0.1788 |
| E15 | 1890.RusselConwell | S | O | 0.124 | 0.748 | -0.1550 | -0.0039 | 0.4742 | 75.7833 | -2.4067 |
| E16 | 1892.FrancesEWHarper | S | O | 0.283 | 0.805 | 0.2143 | -0.0153 | -0.0915 | 39.9448 | 1.3392 |
| E17 | 1901.MarkTwain | S | O | 0.386 | 0.889 | -0.1134 | 0.0263 | 0.0641 | 72.7968 | -0.1564 |
| E18 | 1906.MaryChurch | S | O | 0.375 | 0.852 | 0.1440 | -0.0058 | -0.1499 | 61.9777 | 1.6548 |
| E19 | 1915.AnnaHoward | S | O | 0.134 | 0.775 | -0.2305 | 0.0183 | 0.5420 | 61.0696 | -2.9146 |
| E20 | 1916.CarrieChapman | S | O | 0.252 | 0.794 | 0.2057 | -0.0124 | -0.0303 | 60.8585 | 0.9307 |
| E21 | 1916.HellenKeller | S | O | 0.334 | 0.829 | 0.1983 | -0.0118 | -0.1370 | 45.3861 | 1.6259 |
| E22 | 1918.WoodrowWilson | S | O | 0.279 | 0.818 | 0.0269 | -0.0006 | -0.0112 | 56.7073 | 0.6149 |
| E23 | 1920.CrystalEastman | S | O | 0.314 | 0.848 | 0.0606 | 0.0153 | 0.0675 | 56.1624 | 0.2011 |
| E24 | 1923.JamesMonroe | S | O | 0.354 | 0.849 | -0.0166 | -0.0002 | -0.0816 | 36.3421 | 0.9562 |
| E25 | 1925.MaryReynolds | S | O | 0.198 | 0.802 | -0.1561 | 0.0184 | 0.2870 | 84.2138 | -1.3941 |
| E26 | 1932.MargaretSanger | S | O | 0.343 | 0.847 | -0.0503 | 0.0017 | -0.1212 | 53.0309 | 1.1209 |
| E27 | 1936.EleanorRoosvelt | S | O | 0.232 | 0.830 | -0.2353 | 0.0317 | 0.2744 | 56.2773 | -1.5194 |
| E28 | 1936.KingEdwardVIII | S | O | 0.408 | 0.875 | -0.0954 | 0.0037 | 0.0342 | 65.3628 | 0.0680 |
| E29 | 1941.FranklinDRoosvelt | S | O | 0.455 | 0.881 | -0.0036 | -0.0082 | -0.1906 | 48.2128 | 1.6263 |
| E30 | 1941.HaroldIckes | S | O | 0.294 | 0.822 | 0.0402 | -0.0024 | 0.0248 | 62.8989 | 0.4261 |
| E31 | 1942.MahatmaGandhi | S | O | 0.347 | 0.855 | -0.0215 | 0.0089 | 0.0138 | 63.5228 | 0.3797 |
| E32 | 1944.DwightEisenhower | S | O | 0.577 | 0.925 | -0.0956 | -0.0011 | -0.1486 | 61.7413 | 1.0482 |
| E33 | 1944.GeorgePatton | S | O | 0.359 | 0.886 | -0.0977 | 0.0348 | 0.0323 | 80.0396 | 0.0840 |
| E34 | 1947.GeorgeCMarshall | S | O | 0.362 | 0.842 | 0.1152 | -0.0106 | -0.1892 | 47.2257 | 1.8344 |
| E35 | 1947.HarryTruman | S | O | 0.293 | 0.822 | 0.0353 | -0.0022 | -0.0612 | 46.4032 | 0.9250 |
| E36 | 1950.MargaretChase | S | O | 0.327 | 0.845 | 0.0279 | 0.0068 | -0.0602 | 48.1757 | 0.9075 |
| E37 | 1953.DwightEisenhower | S | O | 0.286 | 0.811 | 0.0689 | -0.0101 | -0.0310 | 59.9117 | 0.7980 |
| E38 | 1953.NelsonMandela | S | O | 0.289 | 0.801 | 0.2887 | -0.0211 | -0.1801 | 43.0789 | 1.9368 |
| E39 | 1957.MartinLutherKing | S | O | 0.159 | 0.780 | -0.1727 | 0.0128 | 0.3431 | 65.5787 | -1.7013 |
| E40 | 1959.RichardFeynman | S | O | 0.160 | 0.785 | -0.1600 | 0.0180 | 0.3665 | 66.4433 | -1.8275 |
| E41 | 1961.01.JohnFKennedy | S | O | 0.348 | 0.852 | 0.0522 | 0.0052 | -0.0371 | 56.0718 | 0.8188 |
| E42 | 1961.04.JohnFKennedy | S | O | 0.353 | 0.845 | 0.1094 | -0.0036 | -0.0980 | 54.3368 | 1.2791 |
| E43 | 1961.05.JohnFKennedy | S | O | 0.233 | 0.799 | 0.1429 | 0.0004 | 0.1528 | 47.2569 | -0.2311 |
| E44 | 1961.11.JohnFKennedy | S | O | 0.465 | 0.892 | 0.0769 | -0.0003 | -0.1246 | 52.3144 | 1.4251 |
| E45 | 1962.09.JohnFKennedy | S | O | 0.309 | 0.827 | 0.0909 | -0.0038 | -0.0497 | 56.0597 | 0.9450 |
| E46 | 1962.10.JohnFKennedy | S | O | 0.293 | 0.829 | 0.0780 | 0.0048 | 0.0344 | 47.9640 | 0.4196 |
| E47 | 1962.12.MalcomX | S | O | 0.095 | 0.757 | -0.3583 | 0.0174 | 0.7080 | 72.7674 | -3.9635 |
| E48 | 1963.06.10.JohnFKennedy | S | O | 0.277 | 0.815 | 0.1203 | -0.0023 | 0.0568 | 56.3615 | 0.3393 |
| E49 | 1963.06.26.JohnFKennedy | S | O | 0.358 | 0.875 | -0.1777 | 0.0241 | -0.0101 | 60.7442 | 0.1277 |
| E50 | 1963.09.20.JohnFKennedy | S | O | 0.273 | 0.804 | 0.1349 | -0.0114 | -0.0233 | 52.3290 | 0.8310 |
| E51 | 1963.MartinLutherKing | S | O | 0.304 | 0.837 | -0.0396 | 0.0077 | 0.0428 | 61.7331 | 0.1824 |
| E52 | 1964.04.MalcomX | S | O | 0.201 | 0.820 | -0.2175 | 0.0349 | 0.3468 | 72.6743 | -1.8614 |
| E53 | 1964.05.LyndonBJohnson | S | O | 0.368 | 0.848 | 0.0199 | -0.0068 | -0.0695 | 60.2291 | 0.9608 |
| E54 | 1964.LadybirdJohnson | S | O | 0.432 | 0.876 | 0.0629 | -0.0040 | -0.1008 | 61.7123 | 1.2446 |
| E55 | 1964.MartinLutherKing | S | O | 0.393 | 0.862 | 0.1191 | -0.0037 | -0.0458 | 50.7437 | 1.0106 |
| E56 | 1964.NelsonMandela | S | O | 0.180 | 0.767 | 0.0760 | -0.0095 | 0.1212 | 53.1394 | -0.1298 |
| E57 | 1965.03.LyndonBJohnson | S | O | 0.234 | 0.805 | -0.0136 | 0.0061 | 0.1485 | 64.1089 | -0.3872 |
| E58 | 1965.04.LyndonBJohnson | S | O | 0.326 | 0.849 | -0.0682 | 0.0108 | 0.0454 | 68.0541 | 0.1031 |
| E59 | 1967.MartinLutherKing | S | O | 0.237 | 0.794 | 0.2044 | -0.0064 | 0.0198 | 55.1386 | 0.6166 |
| E60 | 1968.MartinLutherKing | S | O | 0.196 | 0.793 | -0.1198 | 0.0098 | 0.2791 | 75.3576 | -1.2876 |
| E61 | 1968.RobertFKennedy | S | O | 0.314 | 0.898 | -0.2911 | 0.0649 | 0.2170 | 63.6677 | -1.5177 |
| E62 | 1969.IndiraGhandi | S | O | 0.386 | 0.867 | 0.0340 | 0.0051 | -0.0709 | 65.4832 | 0.9928 |
| E63 | 1969.RichardNixon | S | O | 0.219 | 0.805 | -0.0180 | 0.0123 | 0.1308 | 57.8786 | -0.2912 |
| E64 | 1969.ShirleyChisholm | S | O | 0.395 | 0.867 | 0.0290 | 0.0009 | -0.0465 | 53.8575 | 0.8427 |
| E65 | 1972.JaneFonda | S | O | 0.429 | 0.874 | 0.0462 | -0.0053 | -0.1875 | 44.3886 | 1.7195 |
| E66 | 1972.RichardNixon | S | O | 0.172 | 0.794 | -0.2140 | 0.0213 | 0.3032 | 69.9506 | -1.5466 |
| E67 | 1974.RichardNixon | S | O | 0.274 | 0.833 | -0.1010 | 0.0172 | 0.0518 | 54.6775 | 0.0242 |
| E68 | 1979.MargaretThatcher | S | O | 0.311 | 0.820 | 0.2055 | -0.0114 | -0.1461 | 54.8761 | 1.6676 |
| E69 | 1981.RonaldReagan | S | O | 0.381 | 0.855 | 0.0584 | -0.0052 | -0.0838 | 60.9046 | 1.1204 |
| E70 | 1982.RonaldReagan | S | O | 0.276 | 0.806 | 0.2401 | -0.0108 | -0.0617 | 52.5699 | 1.1728 |
| E71 | 1983.RonaldReagan | S | O | 0.244 | 0.814 | 0.1019 | 0.0105 | 0.1348 | 55.2849 | -0.1673 |
| E72 | 1986.RonaldReagan | S | O | 0.389 | 0.860 | -0.0551 | -0.0034 | -0.0231 | 73.2132 | 0.5214 |
| E73 | 1987.RonaldReagan | S | O | 0.296 | 0.822 | 0.1384 | -0.0035 | -0.0794 | 58.8383 | 1.1760 |
| E74 | 1988.AnnRichards | S | O | 0.278 | 0.826 | 0.0623 | 0.0089 | 0.0951 | 68.2167 | 0.0327 |
| E75 | 1991.GeorgeBush | S | O | 0.329 | 0.844 | 0.0445 | 0.0049 | -0.0115 | 61.5888 | 0.6502 |
| E76 | 1993.MayaAngelou | S | O | 0.392 | 0.835 | -0.0447 | -0.0298 | -0.2499 | 69.3620 | 1.8963 |
| E77 | 1993.SarahBrady | S | O | 0.333 | 0.870 | -0.1386 | 0.0289 | 0.1311 | 70.7339 | -0.5757 |
| E78 | 1993.UrvashiVaid | S | O | 0.315 | 0.840 | -0.0930 | 0.0067 | 0.0114 | 63.9636 | 0.2592 |
| E79 | 1994.NelsonMandela | S | O | 0.384 | 0.848 | 0.0144 | -0.0138 | -0.0936 | 55.1333 | 1.0972 |
| E80 | 1995.ErikaJong | S | O | 0.255 | 0.830 | -0.1092 | 0.0217 | 0.1873 | 56.5715 | -0.7793 |
| E81 | 1995.HillaryClinton | S | O | 0.288 | 0.822 | 0.0232 | 0.0005 | 0.0696 | 54.1626 | 0.1343 |
| E82 | 1997.BillClinton | S | O | 0.320 | 0.845 | -0.0803 | 0.0096 | 0.1107 | 71.2576 | -0.3029 |
| E83 | 1997.EarlOfSpencer | S | O | 0.383 | 0.857 | 0.1062 | -0.0037 | -0.0041 | 57.9058 | 0.7368 |
| E84 | 1997.NancyBirdsall | S | O | 0.279 | 0.833 | -0.0333 | 0.0148 | 0.0783 | 45.2409 | -0.0140 |
| E85 | 1997.PrincessDiana | S | O | 0.343 | 0.850 | 0.0878 | 0.0047 | 0.0174 | 61.3143 | 0.5555 |
| E86 | 1997.QueenElizabethII | S | O | 0.461 | 0.900 | -0.0684 | 0.0087 | -0.0231 | 64.9646 | 0.4532 |
| E87 | 1999.AnitaRoddick | S | O | 0.315 | 0.842 | 0.0446 | 0.0087 | 0.0101 | 60.3981 | 0.5182 |
| E88 | 2000.CondoleezzaRice | S | O | 0.342 | 0.854 | 0.0324 | 0.0093 | -0.0503 | 63.3798 | 0.8583 |
| E89 | 2000.CourtneyLove | S | O | 0.199 | 0.800 | 0.0486 | 0.0157 | 0.1890 | 71.2193 | -0.5628 |
| E90 | 2001.09.11.GeorgeWBush | S | O | 0.442 | 0.881 | 0.0188 | -0.0034 | -0.0727 | 65.8042 | 1.0123 |
| E91 | 2001.09.13.GeorgeWBush | S | O | 0.456 | 0.874 | -0.0139 | -0.0150 | -0.1634 | 57.2094 | 0.9832 |
| E92 | 2001.HalleBerry | S | O | 0.344 | 0.850 | -0.2194 | 0.0042 | 0.0575 | 52.0924 | 1.4434 |
| E93 | 2002.OprahWinfrey | S | O | 0.375 | 0.865 | -0.1652 | 0.0074 | -0.0308 | 83.1871 | -0.3661 |
| E94 | 2003.BethChapman | S | O | 0.381 | 0.876 | -0.0401 | 0.0161 | -0.0435 | 79.1260 | 0.2767 |
| E95 | 2005.SteveJobs | S | O | 0.273 | 0.832 | -0.0163 | 0.0163 | 0.0857 | 62.0435 | 0.6675 |
| E96 | IsaacAsimov.IRobot.Cap2 | N | O | 0.187 | 0.768 | 0.0050 | -0.0110 | 0.1894 | 73.2634 | -0.5956 |
| E97 | IsaacAsimov.IRobot.Cap6 | N | O | 0.154 | 0.760 | -0.0577 | -0.0054 | 0.2709 | 82.1102 | -1.1354 |

# Appendix B

**English texts: Literature Nobel laureates. Readability and Writing Quality Scale comparison**

***G*** *= Genre* [S=Speech: N Novel segment/Story] — ***h*** Entropy [0-1] — ***$J_{1,D}$*** Zipf's deviation

***L*** *= Language* [ E= English : Translation to E] — ***drel*** rel. specific diversity [0-1] — ***RES*** Flesch Read. Easy Score

*d = specific diversity [0-1]* — ***hrel*** rel. entropy [0-1] — ***WQS*** Writing Qualirty Scale

| *Index* | *Text Name* | *G* | *L* | *d* | *h* | *drel* | *hrel* | *$J_{1,D}$* | *RES* | *WSQ* |
|---|---|---|---|---|---|---|---|---|---|---|
| EN1 | 1903.BS.Eng.BjornstjerneBjornson | S | O | 0.347 | 0.850 | 0.0778 | 0.0030 | 0.0487 | 37.7653 | 0.3577 |
| EN2 | 1909.BS.SelmaLagerlof | S | O | 0.273 | 0.825 | -0.0559 | 0.0092 | 0.0186 | 63.3497 | 0.3048 |
| EN3 | 1923.BS.Eng.WilliamButlerYeats | S | O | 0.522 | 0.920 | -0.0569 | 0.0083 | 0.0119 | 51.9322 | 0.2599 |
| EN4 | 1923.NL.Eng.WilliamButlerYeats | S | O | 0.265 | 0.819 | 0.1238 | 0.0073 | 0.0306 | 53.5183 | 0.4847 |
| EN5 | 1930.NL.Eng.SinclairLewis | S | O | 0.282 | 0.799 | 0.3184 | -0.0208 | -0.1360 | 45.8358 | 1.6958 |
| EN6 | 1936.NL.Eng.EugeneOneill | S | O | 0.346 | 0.838 | -0.0396 | -0.0080 | -0.0955 | 49.2763 | 0.9966 |
| EN7 | 1938.BS.PearlBuck | S | O | 0.379 | 0.893 | -0.1964 | 0.0339 | -0.0226 | 61.4490 | 0.1116 |
| EN8 | 1938.NL.PearlBuck | S | O | 0.178 | 0.767 | 0.0088 | -0.0085 | 0.1511 | 56.8045 | -0.3678 |
| EN9 | 1940.05.WinstonChurchill | S | O | 0.415 | 0.873 | -0.0268 | -0.0010 | -0.0931 | 62.9192 | 0.9939 |
| EN10 | 1940.06.A.WinstonChurchill | S | O | 0.284 | 0.823 | 0.1559 | 0.0027 | -0.0642 | 56.0341 | 1.0956 |
| EN11 | 1940.06.B.WinstonChurchill | S | O | 0.243 | 0.802 | 0.0792 | -0.0006 | 0.0320 | 52.6761 | 0.4182 |
| EN12 | 1946.WinstonChurchill | S | O | 0.388 | 0.850 | 0.1080 | -0.0135 | -0.1538 | 51.2463 | 1.6298 |
| EN13 | 1948.BS.Eng.ThomasEliot | S | O | 0.344 | 0.845 | 0.0261 | -0.0003 | -0.0470 | 52.6855 | 0.8334 |
| EN14 | 1949.BS.Eng.WilliamFaulkner | S | O | 0.399 | 0.884 | -0.1028 | 0.0163 | -0.0064 | 64.5715 | 0.2848 |
| EN15 | 1950.NL.Eng.BertrandRussell | S | O | 0.246 | 0.789 | 0.1970 | -0.0147 | -0.0280 | 55.3765 | 0.9045 |
| EN16 | 1954.BS.Eng.ErnestHemingway | S | O | 0.499 | 0.919 | -0.0572 | 0.0154 | -0.0366 | 66.3650 | 0.5480 |
| EN17 | 1962.BS.Eng.JohnSteinbeck | S | O | 0.404 | 0.859 | 0.0473 | -0.0108 | -0.1454 | 55.3406 | 1.4712 |
| EN18 | 1976.BS.Eng.SaulBellow | S | O | 0.504 | 0.912 | -0.0241 | 0.0061 | -0.0981 | 64.1628 | 1.0137 |
| EN19 | 1976.NL.Eng.SaulBellow | S | O | 0.266 | 0.799 | 0.2402 | -0.0139 | -0.0142 | 59.1397 | 0.8841 |
| EN20 | 1983.BS.Eng.WilliamGolding | S | O | 0.545 | 0.913 | 0.0317 | -0.0049 | -0.1896 | 62.6378 | 1.7196 |
| EN21 | 1983.NL.Eng.WilliamGolding | S | O | 0.268 | 0.812 | 0.2085 | -0.0009 | 0.0502 | 64.8120 | 0.4672 |
| EN22 | 1986.BS.Eng.WoleSoyinka | S | O | 0.508 | 0.892 | 0.0515 | -0.0148 | -0.1972 | 43.6372 | 1.8150 |
| EN23 | 1986.NL.Eng.WoleSoyinka | S | O | 0.280 | 0.778 | 0.5240 | -0.0404 | -0.1930 | 43.2831 | 2.1984 |
| EN24 | 1991.BS.Eng.NadineGordimer | S | O | 0.498 | 0.892 | 0.0842 | -0.0116 | -0.1779 | 56.6878 | 1.7757 |
| EN25 | 1991.NL.Eng.NadineGordimer | S | O | 0.286 | 0.802 | 0.2261 | -0.0191 | -0.0878 | 53.4544 | 1.3269 |
| EN26 | 1992.BS.Eng.DerekWalcott | S | O | 0.654 | 0.930 | -0.1846 | -0.0066 | -0.1905 | 24.2505 | 0.8185 |
| EN27 | 1992.NL.Eng.DerekWalcott | S | O | 0.265 | 0.774 | 0.3525 | -0.0381 | -0.1805 | 52.4729 | 1.9739 |
| EN28 | 1993.BS.Eng.ToniMorrison | S | O | 0.546 | 0.912 | 0.0336 | -0.0062 | -0.1078 | 56.9804 | 1.2447 |
| EN29 | 1993.NL.Eng.ToniMorrison | S | O | 0.293 | 0.812 | 0.1663 | -0.0127 | 0.0002 | 65.4967 | 0.7451 |
| EN30 | 1995.BS.Eng.SeamusHeaney | S | O | 0.561 | 0.915 | -0.0220 | -0.0073 | -0.1065 | 53.5405 | 1.0701 |
| EN31 | 1995.NL.Eng.SeamusHeaney | S | O | 0.269 | 0.787 | 0.3492 | -0.0268 | -0.1276 | 49.3848 | 1.6589 |
| EN32 | 2001.BS.Eng.VSNaipaul | S | O | 0.496 | 0.898 | -0.0800 | -0.0049 | -0.1708 | 70.4715 | 1.2791 |
| EN33 | 2001.NL.Eng.VSNaipaul | S | O | 0.194 | 0.787 | -0.0647 | 0.0054 | 0.2345 | 69.6769 | -0.9492 |
| EN34 | 2003.BS.Eng.JMCoetzee | S | O | 0.459 | 0.914 | -0.1630 | 0.0234 | -0.0398 | 77.9339 | 0.2368 |
| EN35 | 2003.NL.Eng.JMCoetzee | S | O | 0.241 | 0.793 | 0.0472 | -0.0085 | -0.0467 | 65.3898 | 0.8479 |
| EN36 | 2005.NL.Eng.HaroldPinter | S | O | 0.255 | 0.802 | 0.1998 | -0.0058 | -0.0002 | 65.9238 | 0.7471 |
| EN37 | 2007.NL.Eng.DorisLessing | S | O | 0.212 | 0.792 | -0.0010 | 0.0029 | 0.0405 | 69.1879 | 0.2667 |
| EN38 | ErnestHemingway.TheOldManAndTheSea.Part1 | N | O | 0.121 | 0.744 | -0.2234 | -0.0070 | 0.3077 | 82.9544 | -1.4910 |
| EN39 | ErnestHemingway.TheOldManAndTheSea.Part2 | N | O | 0.110 | 0.738 | -0.2750 | -0.0076 | 0.3865 | 82.6875 | -1.9928 |
| EN40 | ErnestHemingway.TheSunAlsoRises.Book1 | N | O | 0.091 | 0.705 | -0.3231 | -0.0333 | 0.4104 | 94.1593 | -2.1292 |
| ENT1 | 1945.BS.Eng.GabrielaMistral | S | T | 0.530 | 0.883 | 0.0042 | -0.0305 | -0.2446 | 80.0260 | 1.9787 |
| ENT2 | 1967.BS.Eng.MiguelAngelAsturias | S | T | 0.419 | 0.849 | 0.1159 | -0.0264 | -0.2004 | 40.2646 | 1.9431 |
| ENT3 | 1967.NL.Eng.MiguelAngelAsturias | S | T | 0.294 | 0.790 | 0.3196 | -0.0347 | -0.2295 | 57.9760 | 2.2724 |
| ENT4 | 1971.BS.Eng.PabloNeruda | S | T | 0.416 | 0.870 | -0.1283 | -0.0040 | -0.1067 | 47.9302 | 0.8021 |
| ENT5 | 1971.NL.Eng.PabloNeruda | S | T | 0.280 | 0.811 | 0.1733 | -0.0069 | -0.0349 | 52.8923 | 0.9452 |
| ENT6 | 1977.NL.Eng.VicenteAleixandre | S | T | 0.323 | 0.827 | 0.1671 | -0.0098 | -0.0723 | 49.7503 | 1.1941 |
| ENT7 | 1982.NL.Eng.GabrielGarciaMarquez | S | T | 0.391 | 0.840 | 0.3202 | -0.0242 | -0.2080 | 56.0816 | 2.2687 |
| ENT8 | 1989.BS.Eng.CamioJoseCela | S | T | 0.489 | 0.892 | -0.0006 | -0.0093 | -0.1957 | 44.9517 | 1.6645 |
| ENT9 | 1989.NL.Eng.CamioJoseCela | S | T | 0.265 | 0.806 | 0.2453 | -0.0063 | -0.0364 | 52.4467 | 1.0133 |
| ENT10 | 1990.BS.Eng.OctavioPaz | S | T | 0.472 | 0.869 | 0.0693 | -0.0264 | -0.1903 | 45.4568 | 1.8122 |
| ENT11 | 1990.NL.Eng.OctavioPaz | S | T | 0.272 | 0.787 | 0.2697 | -0.0280 | -0.1224 | 58.2610 | 1.5628 |
| ENT12 | 2010.BS.Eng.MarioVargasLlosa | S | T | 0.451 | 0.891 | -0.0860 | 0.0035 | -0.0722 | 57.6237 | 0.6978 |
| ENT13 | 2010.NL.Eng.MarioVargasLlosa | S | T | 0.288 | 0.774 | 0.4434 | -0.0478 | -0.2736 | 52.6690 | 2.6334 |

## Appendix C

**Spanish texts: Literature Non-Nobel laureates. Readability and Writing Quality Scale comparison**

| | | |
|---|---|---|
| ***G** = Genre* [S=Speech: N Novel segment/Story] | ***h*** Entropy [0-1] | ***J1,D*** Zipf's deviation |
| ***L** = Language* [ E= English : Translation to E] | ***drel*** rel. specific diversity [0-1] | ***IPSZ*** Szigritsz Perpicuity index |
| *d = specific diversity [0-1]* | ***hrel*** rel. entropy [0-1] | ***WQS*** Writing Quality Scale |

| *Index* | *Text Name* | *G* | *L* | *d* | *h* | *drel* | *hrel* | *J1,D* | *IPSZ* | *WSQ* |
|---|---|---|---|---|---|---|---|---|---|---|
| ST1 | 1755.PatrickHenry | S | T | 0.482 | 0.910 | -0.0895 | 0.0204 | -0.1309 | 77.192 | -0.2454 |
| ST2 | 1863.AbrahamLincoln | S | T | 0.498 | 0.923 | -0.0655 | 0.0268 | -0.1147 | 63.068 | -0.2007 |
| ST3 | 1873.SusanBAnthony | S | T | 0.399 | 0.870 | -0.1162 | 0.0178 | -0.0055 | 66.189 | -1.0279 |
| ST4 | 1899.Vladimir Lenin | S | T | 0.335 | 0.817 | -0.0038 | -0.0047 | -0.1405 | 50.814 | 0.3061 |
| ST5 | 1918.WoodrowWilson | S | T | 0.564 | 0.909 | 0.0565 | -0.0111 | -0.1577 | 61.409 | 0.7354 |
| ST6 | 1919.Georges Clemenceau | S | T | 0.603 | 0.928 | 0.0285 | -0.0040 | -0.1792 | 73.798 | 0.6815 |
| ST7 | 1919.Lloyd George | S | T | 0.681 | 0.950 | 0.0423 | 0.0049 | -0.1906 | 46.434 | 0.8074 |
| ST8 | 1921.MarieCurie.Esp | S | T | 0.425 | 0.895 | -0.0722 | 0.0306 | -0.0097 | 65.611 | -0.7744 |
| ST9 | 1934.Adolf Hitler | S | T | 0.470 | 0.893 | -0.0903 | 0.0084 | -0.1460 | 70.292 | -0.1612 |
| ST10 | 1938.Leon Trotsky | S | T | 0.407 | 0.860 | 0.0319 | 0.0040 | -0.0634 | 59.841 | 0.1045 |
| ST11 | 1938.Neville Chamberlain | S | T | 0.473 | 0.883 | 0.0675 | -0.0032 | -0.1313 | 59.572 | 0.6544 |
| ST12 | 1940.Benito Mussolini | S | T | 0.459 | 0.868 | 0.0733 | -0.0121 | -0.2212 | 62.692 | 1.1519 |
| ST13 | 1940.Charles de Gaulle | S | T | 0.566 | 0.928 | -0.1566 | 0.0075 | -0.0981 | 64.763 | -0.7726 |
| ST14 | 1941.Franklin Roosevelt | S | T | 0.564 | 0.922 | 0.0357 | 0.0010 | -0.1514 | 55.837 | 0.5753 |
| ST15 | 1941.Joseph Stalin | S | T | 0.388 | 0.878 | -0.0530 | 0.0307 | -0.0206 | 56.737 | -0.6124 |
| ST16 | 1942.08.Mahatma Gandhi | S | T | 0.334 | 0.825 | 0.0684 | 0.0045 | -0.0251 | 78.587 | 0.1130 |
| ST17 | 1943.Heinrich Himmler | S | T | 0.526 | 0.919 | 0.0203 | 0.0114 | -0.1970 | 70.493 | 0.7103 |
| ST18 | 1943.Joseph Goebbels | S | T | 0.353 | 0.864 | -0.0732 | 0.0335 | 0.1007 | 70.605 | -1.3435 |
| ST19 | 1945.Harry Truman | S | T | 0.410 | 0.869 | -0.0312 | 0.0110 | -0.1099 | 65.903 | -0.0176 |
| ST20 | 1945.Hirohito | S | T | 0.460 | 0.868 | 0.0847 | -0.0119 | -0.1625 | 52.164 | 0.9178 |
| ST21 | 1947.George Marshall | S | T | 0.435 | 0.867 | 0.0228 | -0.0020 | -0.1276 | 54.028 | 0.3860 |
| ST22 | 1948.David Ben Gurion | S | T | 0.354 | 0.826 | -0.0695 | -0.0050 | -0.1490 | 40.487 | -0.0160 |
| ST23 | 1950.Robert Schuman | S | T | 0.386 | 0.840 | -0.0271 | -0.0058 | -0.2110 | 46.696 | 0.5350 |
| ST24 | 1950.William Faulkner | S | T | 0.437 | 0.895 | -0.0575 | 0.0243 | -0.0062 | 74.087 | -0.7044 |
| ST25 | 1953.Dwight D Eisenhower | S | T | 0.359 | 0.847 | 0.0395 | 0.0140 | -0.1146 | 53.995 | 0.3966 |
| ST26 | 1956.Gamar Abdel Nasser | S | T | 0.402 | 0.868 | -0.0300 | 0.0141 | -0.0501 | 58.889 | -0.3180 |
| ST27 | 1959.Nikita Kruschev | S | T | 0.490 | 0.889 | -0.0141 | -0.0041 | -0.1613 | 60.915 | 0.3535 |
| ST28 | 1961.J F Kennedy | S | T | 0.373 | 0.838 | 0.0613 | -0.0017 | -0.1359 | 60.538 | 0.6420 |
| ST29 | 1961.Nelson Mandela | S | T | 0.257 | 0.778 | -0.0152 | -0.0029 | -0.0317 | 64.026 | -0.3113 |
| ST30 | 1962.J F Kennedy | S | T | 0.502 | 0.905 | -0.0489 | 0.0069 | -0.1678 | 75.548 | 0.1820 |
| ST31 | 1963.J F Kennedy | S | T | 0.393 | 0.891 | -0.1087 | 0.0408 | 0.0699 | 69.876 | -1.3920 |
| ST32 | 1963.Martin Luther King Jr | S | T | 0.331 | 0.828 | -0.0398 | 0.0090 | -0.1139 | 68.380 | -0.0434 |
| ST33 | 1964.Malcom X | S | T | 0.390 | 0.877 | -0.0635 | 0.0288 | 0.0284 | 75.042 | -0.9175 |
| ST34 | 1964.Nelson Mandela | S | T | 0.257 | 0.778 | -0.0155 | -0.0028 | -0.0318 | 63.910 | -0.3129 |
| ST35 | 1964.Ronald Reagan | S | T | 0.424 | 0.875 | 0.0854 | 0.0112 | -0.0485 | 59.968 | 0.3193 |
| ST36 | 1967.Martin Luther King | S | T | 0.259 | 0.786 | 0.0801 | 0.0035 | -0.0666 | 62.289 | 0.3892 |
| ST37 | 1969.Richard Nixon | S | T | 0.267 | 0.800 | -0.0202 | 0.0138 | -0.0257 | 61.460 | -0.3864 |
| ST38 | 1974.Richard Nixon | S | T | 0.408 | 0.879 | -0.0459 | 0.0220 | 0.0089 | 62.354 | -0.7139 |
| ST39 | 1979.Ayatolá Jomeini | S | T | 0.496 | 0.918 | -0.1114 | 0.0225 | -0.0487 | 73.056 | -0.7866 |
| ST40 | 1982.Margaret Thatcher | S | T | 0.413 | 0.895 | -0.0883 | 0.0358 | -0.0031 | 66.736 | -0.9030 |
| ST41 | 1984.Ronald Reagan | S | T | 0.429 | 0.864 | 0.0208 | -0.0030 | -0.0730 | 71.437 | 0.0983 |
| ST42 | 1986.Ronald Reagan | S | T | 0.443 | 0.879 | 0.0330 | 0.0055 | -0.1528 | 68.187 | 0.5628 |
| ST43 | 1987.Ronald Reagan | S | T | 0.323 | 0.816 | 0.0842 | 0.0015 | -0.1437 | 64.967 | 0.8057 |
| ST44 | 1988.Gorbachov | S | T | 0.409 | 0.859 | 0.0365 | 0.0015 | -0.0930 | 54.310 | 0.2825 |
| ST45 | 1990.George H. W. Bush | S | T | 0.411 | 0.881 | -0.0667 | 0.0222 | -0.1144 | 62.149 | -0.2039 |
| ST46 | 1991.Boris Yeltsin | S | T | 0.470 | 0.889 | -0.0203 | 0.0045 | -0.1596 | 55.557 | 0.3018 |
| ST47 | 1991.Gorbachov | S | T | 0.640 | 0.936 | 0.0752 | -0.0035 | -0.1333 | 68.773 | 0.7077 |
| ST48 | 1992.Severn Suzuki | S | T | 0.403 | 0.869 | 0.0161 | 0.0144 | -0.1567 | 75.017 | 0.4798 |
| ST49 | 1993.Bill Clinton | S | T | 0.350 | 0.827 | 0.0508 | -0.0020 | -0.0983 | 66.216 | 0.3929 |
| ST50 | 1999.Elie Wiesel | S | T | 0.407 | 0.854 | -0.0271 | -0.0022 | -0.2175 | 69.381 | 0.5647 |
| ST51 | 2001.George W. Bush | S | T | 0.509 | 0.905 | -0.0196 | 0.0045 | -0.1551 | 59.143 | 0.2824 |
| ST52 | 2001.Osama Bin Laden | S | T | 0.473 | 0.891 | -0.0208 | 0.0052 | -0.1053 | 74.907 | 0.0228 |
| ST53 | 2002.A.George W. Bush | S | T | 0.459 | 0.887 | 0.0158 | 0.0067 | -0.0949 | 65.758 | 0.1716 |
| ST54 | 2002.Barack Hussein Obama | S | T | 0.386 | 0.840 | -0.0313 | -0.0060 | -0.0680 | 67.871 | -0.2142 |
| ST55 | 2003.B.George W. Bush | S | T | 0.420 | 0.886 | -0.0811 | 0.0235 | -0.0557 | 65.029 | -0.5837 |
| ST56 | 2003.George W. Bush | S | T | 0.475 | 0.879 | 0.1121 | -0.0077 | -0.1636 | 57.980 | 1.0715 |
| ST57 | 2005.Gerhard Schroeder | S | T | 0.361 | 0.843 | 0.0168 | 0.0084 | -0.0765 | 64.530 | 0.0824 |
| ST58 | 2005.Steve Jobs | S | T | 0.330 | 0.831 | 0.0484 | 0.0123 | -0.0615 | 75.390 | 0.1780 |
| ST59 | 2006.Dianne Feinstein | S | T | 0.349 | 0.841 | -0.0241 | 0.0125 | -0.0883 | 70.932 | -0.0896 |
| ST60 | 2007.Al Gore | S | T | 0.440 | 0.859 | 0.1891 | -0.0125 | -0.2285 | 64.044 | 1.8339 |
| ST61 | 2008.Barack Hussein Obama | S | T | 0.515 | 0.897 | -0.0320 | -0.0056 | -0.1204 | 76.954 | 0.0477 |
| ST62 | 2008.Randy Paush | S | T | 0.343 | 0.847 | 0.0061 | 0.0216 | -0.0798 | 80.083 | 0.0262 |
| ST63 | 2009.Barack Hussein Obama | S | T | 0.345 | 0.817 | 0.1298 | -0.0095 | -0.0886 | 64.936 | 0.7909 |
| ST64 | 2010.Hillary Clinton | S | T | 0.343 | 0.831 | 0.0800 | 0.0063 | -0.1075 | 49.720 | 0.5936 |
| ST65 | IsaacAsimov.YoRobot.Cap2 | N | T | 0.230 | 0.767 | -0.0228 | 0.0003 | -0.0198 | 84.229 | -1.2116 |
| ST66 | IsaacAsimov.YoRobot.Cap6 | N | T | 0.195 | 0.754 | -0.0778 | 0.0049 | 0.0754 | 81.050 | -0.8770 |
| S1 | 1805.Simón Bolívar | S | O | 0.498 | 0.878 | 0.0356 | -0.0184 | -0.1901 | 68.058 | 0.7910 |
| S2 | 1813.Simón Bolívar | S | O | 0.449 | 0.864 | 0.0510 | -0.0116 | -0.0863 | 56.710 | 0.3427 |
| S3 | 1819.Simón Bolívar | S | O | 0.229 | 0.751 | 0.0621 | -0.0153 | -0.0302 | 59.935 | 0.1238 |
| S4 | 1830.Simón Bolívar | S | O | 0.602 | 0.930 | 0.0171 | -0.0020 | -0.1123 | 68.505 | 0.2763 |
| S5 | 1868.CarlosMCespedes | S | O | 0.406 | 0.836 | 0.1245 | -0.0194 | -0.2141 | 46.264 | 1.4088 |
| S6 | 1912.Emiliano Zapata | S | O | 0.361 | 0.811 | 0.1556 | -0.0228 | -0.2512 | 45.508 | 1.7732 |
| S7 | 1917.Emiliano Zapata | S | O | 0.403 | 0.826 | 0.1480 | -0.0289 | -0.2595 | 44.732 | 1.7796 |
| S8 | 1918.Emiliano Zapata | S | O | 0.412 | 0.830 | 0.1394 | -0.0292 | -0.2428 | 42.040 | 1.6473 |
| S9 | 1931.Manuel Azaña | S | O | 0.512 | 0.906 | -0.0467 | 0.0037 | -0.1526 | 61.079 | 0.1198 |
| S10 | 1933.JAntonioPrimoDeRivera | S | O | 0.305 | 0.803 | 0.0275 | -0.0025 | -0.1575 | 64.229 | 0.5642 |
| S11 | 1936.Dolores Ibarruri | S | O | 0.359 | 0.864 | -0.2237 | 0.0311 | 0.1382 | 66.835 | -2.3690 |
| S12 | 1936.José Buenaventura Durruti | S | O | 0.442 | 0.877 | 0.0165 | 0.0049 | -0.0492 | 69.053 | -0.0539 |
| S13 | 1938.Dolores Ibarruri | S | O | 0.411 | 0.846 | -0.0276 | -0.0124 | -0.2177 | 58.630 | 0.5726 |
| S14 | 1945.Juan Domingo Perón | S | O | 0.391 | 0.865 | 0.0007 | 0.0165 | -0.0695 | 67.165 | -0.0511 |
| S15 | 1946.Jorge Eliécer Gaitán | S | O | 0.278 | 0.811 | -0.0368 | 0.0192 | 0.0390 | 52.904 | -0.8132 |
| S16 | 1952.Eva Perón | S | O | 0.306 | 0.839 | -0.2049 | 0.0325 | 0.0302 | 66.053 | -1.7178 |
| S17 | 1959.Fidel Castro | S | O | 0.295 | 0.810 | -0.0295 | 0.0095 | -0.0116 | 66.794 | 0.3802 |
| S18 | 1959.Fulgencio Batista | S | O | 0.682 | 0.947 | -0.0703 | 0.0014 | -0.1683 | 46.517 | 0.0703 |
| S19 | 1964.Ernesto Che Guevara | S | O | 0.266 | 0.779 | 0.1002 | -0.0072 | -0.1346 | 64.026 | 0.8579 |
| S20 | 1967.Ernesto Che Guevara | S | O | 0.289 | 0.788 | 0.1351 | -0.0095 | -0.1203 | 49.783 | 0.9811 |
| S21 | 1967.Fidel Castro | S | O | 0.223 | 0.788 | -0.1367 | 0.0240 | 0.0186 | 56.669 | -1.2705 |
| S22 | 1970.Salvador Allende | S | O | 0.385 | 0.834 | 0.1352 | -0.0119 | -0.1467 | 56.961 | 1.1185 |
| S23 | 1972.Salvador Allende | S | O | 0.253 | 0.766 | 0.1358 | -0.0129 | -0.1413 | 45.267 | 1.0952 |
| S24 | 1973.Augusto Pinochet | S | O | 0.314 | 0.797 | 0.1354 | -0.0142 | -0.1215 | 51.864 | 0.9935 |
| S25 | 1973.Bando Nro 5 | S | O | 0.457 | 0.859 | 0.0000 | 0.0000 | -0.2092 | 48.134 | 1.1957 |
| S26 | 1973.Salvador Allende | S | O | 0.449 | 0.868 | 0.0353 | -0.0075 | -0.1743 | 65.375 | 0.6981 |
| S27 | 1976.Jorge Videla | S | O | 0.437 | 0.875 | -0.0277 | 0.0043 | -0.1833 | 56.853 | 0.3806 |
| S28 | 1978.Juan Carlos I | S | O | 0.422 | 0.848 | 0.0586 | -0.0153 | -0.1884 | 45.654 | 0.9068 |
| S29 | 1979.Adolfo Suárez | S | O | 0.212 | 0.751 | 0.0198 | -0.0071 | -0.1023 | 42.830 | 0.2457 |
| S30 | 1979.Fidel Castro | S | O | 0.208 | 0.743 | -0.0071 | -0.0122 | -0.0490 | 50.503 | -0.1691 |
| S31 | 1981.Adolfo Suárez | S | O | 0.312 | 0.842 | -0.1529 | 0.0327 | 0.0881 | 61.705 | -1.7222 |
| S32 | 1981.Roberto Eduardo Viola | S | O | 0.337 | 0.799 | 0.1887 | -0.0232 | -0.1744 | 48.864 | 1.5685 |
| S33 | 1982.Felipe González | S | O | 0.276 | 0.782 | 0.1150 | -0.0086 | -0.0994 | 48.228 | 0.7628 |
| S34 | 1982.Leopoldo Galtieri | S | O | 0.639 | 0.934 | -0.0535 | -0.0057 | -0.2430 | 45.919 | -4.4241 |
| S35 | 1983.Raúl Alfonsín | S | O | 0.295 | 0.805 | 0.0038 | 0.0043 | 0.0097 | 58.433 | 0.1707 |
| S36 | 1989.Carlos Saúl Menem | S | O | 0.337 | 0.845 | -0.1103 | 0.0231 | 0.0081 | 41.623 | -1.0698 |
| S37 | 1992.Rafael Caldera | S | O | 0.332 | 0.810 | 0.0546 | -0.0098 | -0.1498 | 46.915 | 0.6837 |
| S38 | 1996.Jose María Aznar | S | O | 0.273 | 0.782 | 0.0330 | -0.0079 | -0.1035 | 62.146 | 0.3260 |
| S39 | 1999.Hugo Chavez | S | O | 0.191 | 0.760 | -0.0880 | 0.0129 | -0.0507 | 58.903 | -0.6367 |
| S40 | 2000.Vicente Fox | S | O | 0.269 | 0.778 | 0.1217 | -0.0098 | -0.0660 | 63.589 | 0.6314 |
| S41 | 2001.Fernando de la Rúa | S | O | 0.386 | 0.853 | 0.0044 | 0.0066 | -0.0422 | 48.038 | -0.1589 |
| S42 | 2004.Pilar Manjón | S | O | 0.565 | 0.917 | -0.0368 | -0.0032 | -0.2086 | 66.429 | 0.4663 |
| S43 | 2005.Daniel Ortega | S | O | 0.200 | 0.779 | -0.1637 | 0.0275 | 0.1246 | 54.243 | -1.9628 |
| S44 | 2006.Alvaro Uribe | S | O | 0.341 | 0.776 | 0.2560 | -0.0477 | -0.2444 | 57.430 | 2.3228 |
| S45 | 2006.Evo Morales | S | O | 0.262 | 0.812 | -0.1013 | 0.0277 | -0.1543 | 53.388 | -0.1996 |
| S46 | 2006.Gastón Acurio | S | O | 0.293 | 0.803 | 0.0693 | 0.0034 | -0.1481 | 57.135 | 0.7430 |
| S47 | 2006.Hugo Chavez | S | O | 0.283 | 0.808 | -0.0346 | 0.0136 | -0.1495 | 54.065 | 0.1615 |
| S48 | 2007.Cristina Kirchner | S | O | 0.245 | 0.795 | -0.0739 | 0.0199 | 0.0469 | 46.475 | -1.0604 |
| S49 | 2007.Daniel Ortega | S | O | 0.254 | 0.805 | -0.1312 | 0.0252 | -0.0824 | 59.970 | -0.7284 |
| S50 | 2008.J. L. Rodriguez Zapatero | S | O | 0.454 | 0.886 | -0.0616 | 0.0078 | -0.0399 | 74.018 | -0.5390 |
| S51 | 2008.Julio Cobos | S | O | 0.493 | 0.907 | -0.0954 | 0.0130 | -0.0492 | 44.405 | -0.6854 |
| S52 | 2010.Raúl Castro | S | O | 0.558 | 0.912 | 0.0048 | -0.0069 | -0.2289 | 65.089 | 0.8051 |
| S53 | 2010.Sebastian Piñera Echenique | S | O | 0.400 | 0.890 | -0.1808 | 0.0363 | -0.0374 | 75.856 | -1.2446 |
| S54 | JorgeLuisBorges.ElCongreso | N | O | 0.289 | 0.774 | 0.1727 | -0.0242 | -0.1398 | 75.849 | 1.3048 |
| S55 | JorgeLuisBorges.ElMuerto | N | O | 0.357 | 0.814 | 0.0857 | -0.0185 | -0.1740 | 73.621 | 0.9880 |
| S56 | JorgeLuisBorges.ElSur | N | O | 0.345 | 0.800 | 0.1213 | -0.0263 | -0.1925 | 70.208 | 1.2885 |
| S57 | JorgeLuisBorges.LasRuinasCirculares | N | O | 0.368 | 0.826 | 0.1363 | -0.0121 | -0.1380 | 72.796 | -0.4179 |

# Appendix D

**Spanish texts: Literature Nobel laureates. Readability and Writing Quality Scale comparison**

| | | |
|---|---|---|
| ***G** = Genre* [S=Speech: N Novel segment/Story] | ***h*** Entropy [0-1] | ***J<sub></sub>*** |
| ***L** = Language* [ E= English : Translation to E] | ***drel*** rel. specific diversity [0-1] | ***IPSZ*** Szigritsz Perpicuity index |
| *d = specific diversity [0-1]* | ***hrel*** rel. entropy [0-1] | ***WQS*** Writing Quality Scale |

| *Index* | *Text Name* | *G* | *L* | *d* | *h* | *drel* | *hrel* | $J_{1,D}$ | *IPSZ* | *WSQ* |
|---|---|---|---|---|---|---|---|---|---|---|
| SN1 | 1967.BS.Esp.MiguelAngelAsturias | S | O | 0.422 | 0.845 | 0.0074 | -0.018 | -0.2033 | 72.818 | 0.1601 |
| SN2 | 1967.NL.Esp.MiguelAngelAsturias | S | O | 0.313 | 0.787 | 0.1743 | -0.023 | -0.1844 | 58.872 | 0.1601 |
| SN3 | 1971.BS.Esp.PabloNeruda | S | O | 0.447 | 0.859 | -0.068 | -0.016 | -0.1928 | 63.482 | 1.2979 |
| SN4 | 1971.Pablo Neruda | S | O | 0.35 | 0.806 | 0.2243 | -0.023 | -0.2226 | 59.511 | 0.1969 |
| SN5 | 1977.BS.Esp.VicenteAleixandre | S | O | 0.568 | 0.917 | 0.005 | -0.005 | -0.2092 | 64.889 | 1.4512 |
| SN6 | 1977.NL.Esp.VicenteAleixandre | S | O | 0.361 | 0.818 | 0.1315 | -0.016 | -0.2534 | 67.929 | -1.7016 |
| SN7 | 1982.BS.Esp.GabrielGarciaMarquez | S | O | 0.481 | 0.876 | 0.0313 | -0.013 | -0.1567 | 58.857 | -2.5459 |
| SN8 | 1982.Gabriel García Márquez | S | O | 0.409 | 0.831 | 0.2403 | -0.026 | -0.2419 | 54.692 | -1.0469 |
| SN9 | 1987.Camilo José Cela | S | O | 0.390 | 0.830 | 0.1061 | -0.0181 | -0.2049 | 65.607 | 0.7009 |
| SN10 | 1989.NL.Esp.CamiloJoseCela | S | O | 0.287 | 0.777 | 0.1453 | -0.02 | -0.1478 | 61.068 | 1.5391 |
| SN11 | 1990.BS.Esp.OctavioPaz | S | O | 0.463 | 0.878 | 0.0344 | -0.004 | -0.1311 | 54.83 | 0.2251 |
| SN12 | 1990.NL.Esp.OctavioPaz | S | O | 0.302 | 0.788 | 0.1291 | -0.016 | -0.076 | 64.924 | 2.0110 |
| SN13 | 2010.BS.Esp.MarioVargasLlosa | S | O | 0.481 | 0.888 | -0.02 | -0.001 | -0.2166 | 69.764 | 0.7044 |
| SN14 | 2010.NL.Esp.MarioVargasLlosa | S | O | 0.315 | 0.763 | 0.294 | -0.048 | -0.3184 | 63.788 | 1.6440 |
| SN15 | CamiloJoseCela.LaColmena.Cap1 | S | O | 0.177 | 0.736 | -0.085 | -0.003 | 0.0027 | 49.293 | 0.5922 |
| SN16 | CamiloJoseCela.LaColmena.Cap2 | S | O | 0.191 | 0.741 | -0.043 | -0.005 | -0.0057 | 88.945 | 2.2009 |
| SN17 | CamiloJoseCela.LaColmena.Cap6 | S | O | 0.308 | 0.798 | 0.0719 | -0.009 | -0.2226 | 88.846 | 1.2583 |
| SN18 | CamiloJoseCela.LaColmena.Notas4Ediciones | S | O | 0.367 | 0.829 | 0.0458 | -0.008 | -0.1715 | 82.21 | 1.1881 |
| SN19 | GabrielGMarquez.CronMuerteAnunciada.Cap1y2 | N | O | 0.21 | 0.754 | -0.002 | -0.003 | 0.0802 | 71.553 | 0.4698 |
| SN20 | GabrielGMarquez.CronMuerteAnunciada.Cap3y4 | N | O | 0.218 | 0.754 | 0.0364 | -0.006 | 0.0578 | 70.095 | 0.7301 |
| SN21 | GabrielGMarquez.CronMuerteAnunciada.Last | N | O | 0.235 | 0.774 | -0.044 | 0.0048 | 0.0882 | 75.775 | 0.5960 |
| SN22 | GabrielGMarquez.DicursoCartagena | S | O | 0.401 | 0.844 | 0.1097 | -0.009 | -0.1753 | 67.05 | 2.9105 |
| SN23 | GabrielGMarquez.MejorOficioDelMundo | S | O | 0.359 | 0.808 | 0.1874 | -0.025 | -0.1859 | 55.767 | 0.7425 |
| SN24 | MarioVargasLlosa.DiscursoBuenosAires | S | O | 0.391 | 0.819 | 0.1713 | -0.029 | -0.2463 | 50.209 | 1.1190 |
| SN25 | MiguelAAsturias.SrPresidente.Parte1.Cap1y2 | N | O | 0.292 | 0.786 | 0.0627 | -0.013 | -0.1428 | 76.649 | 1.6215 |
| SN26 | OctavioPaz.DiscursoZacatecas | S | O | 0.318 | 0.81 | -0.02 | -0.002 | -0.1014 | 68.929 | 1.8427 |
| SN27 | OctavioPaz.LaberintoSoledad.Part3 | N | O | 0.261 | 0.757 | 0.0744 | -0.027 | -0.1427 | 70.767 | 0.0171 |
| SNT1 | 1940.B.Winston Churchill | S | T | 0.529 | 0.892 | -0.3179 | -0.0166 | -0.1364 | 72.469 | -0.5983 |
| SNT2 | 1940.Winston Churchill | S | T | 0.494 | 0.900 | -0.0125 | 0.0058 | -0.1234 | 57.863 | 1.1487 |
| SNT3 | 1998.José Saramago | S | T | 0.285 | 0.781 | 0.1351 | -0.0141 | -0.1817 | 67.259 | -0.8087 |
| SNT4 | 2003.José Saramago | S | T | 0.397 | 0.849 | 0.0290 | -0.0028 | -0.0835 | 83.911 | -0.4755 |
| SNT5 | ErnestHemingway.ElViejoYElMar.Part1 | N | T | 0.179 | 0.751 | -0.1281 | 0.0106 | 0.1155 | 85.557 | -1.0914 |
| SNT6 | ErnestHemingway.ElViejoYElMar.Part2 | N | T | 0.157 | 0.743 | -0.2150 | 0.0147 | 0.1858 | 90.348 | 1.0804 |
| SNT7 | ErnestHemingway.Fiesta.Libro1 | N | T | 0.174 | 0.733 | -0.1019 | -0.0046 | 0.0184 | 77.504 | 0.6961 |
| SNT8 | JoseSaramago.Valencia | S | T | 0.303 | 0.786 | 0.0626 | -0.0193 | -0.2904 | 49.823 | 0.7743 |

## Appendix E. Writing Quality Scale for English speeches

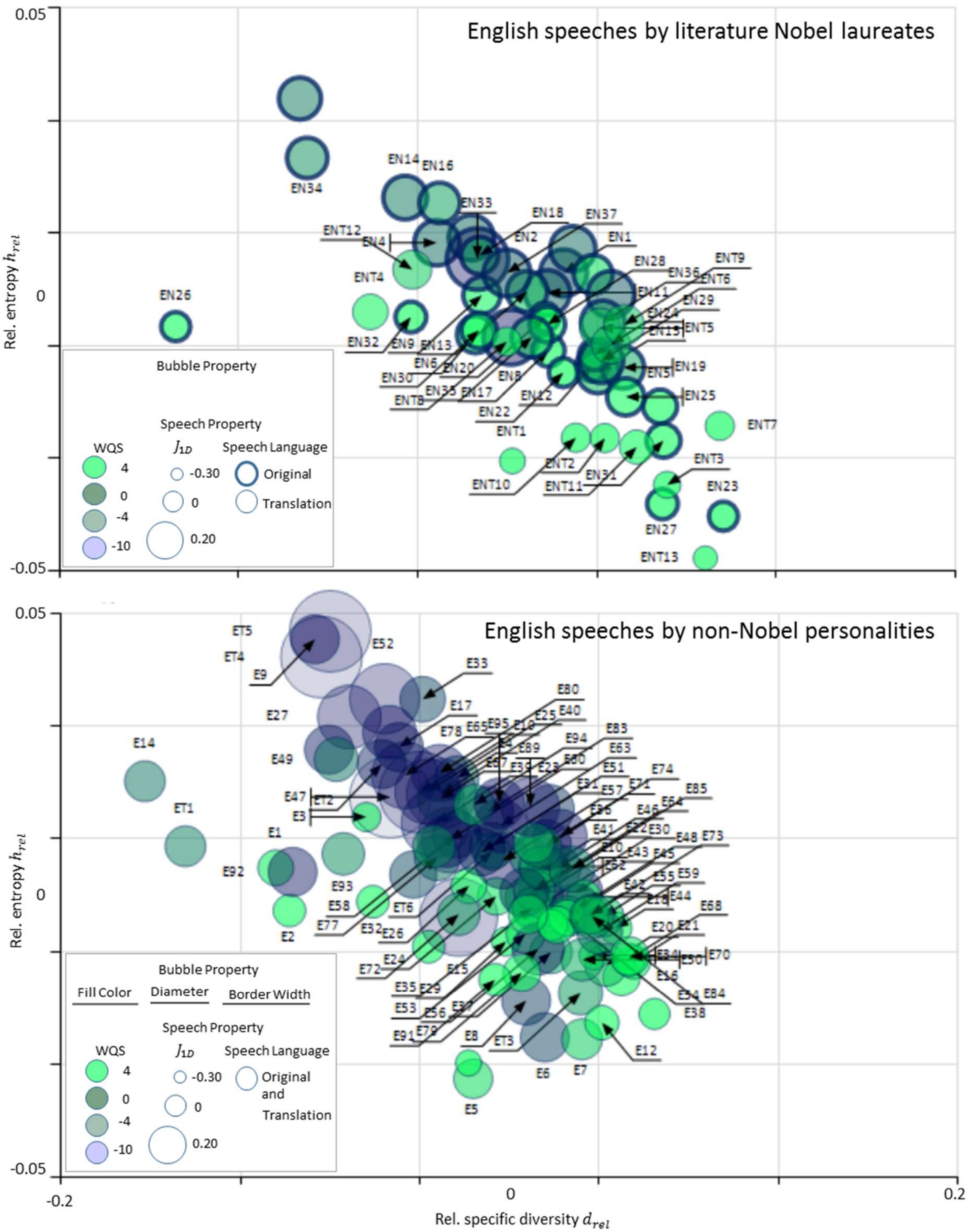

# Appendix F. Writing Quality Scale for Spanish speeches

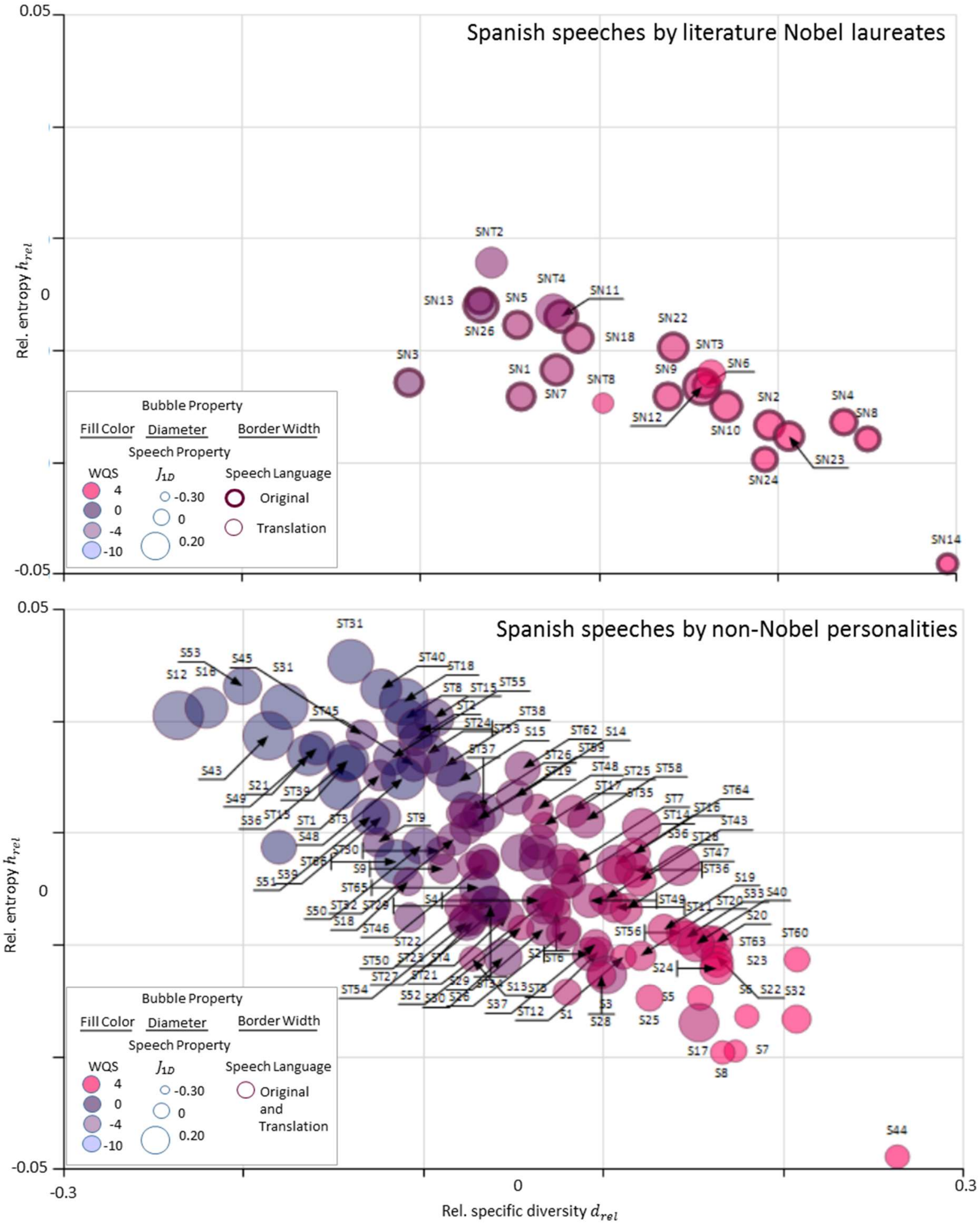